\documentclass[11pt]{article}
\usepackage{ruc-tech-report}
\usepackage{amsmath,amssymb,amsfonts}
\usepackage{booktabs,multirow,array,tabularx}
\usepackage{rotating,pifont,float,subcaption}
\usepackage[export]{adjustbox}
\usepackage{hyperref}
\hypersetup{colorlinks=true,linkcolor=rucred,citecolor=rucred,urlcolor=rucred}
\setkeys{Gin}{max width=\linewidth,max totalheight=0.78\textheight,keepaspectratio}

\newcommand{\authororcid}[2]{{\hypersetup{hidelinks}\href{https://orcid.org/#1}{#2}}}
\providecommand{\Description}[2][]{}
\newcommand{\papertitle}{Beyond Static Charts: Benchmarking Interactive Chart-to-Code Generation for MLLMs}
\newcommand{\paperauthors}{\authororcid{0009-0001-0312-8101}{Yuhan Bian}$^{1,*}$\hspace{0.6em}\authororcid{0009-0008-8246-7056}{Haoyu Liu}$^{1,*}$\hspace{0.6em}\authororcid{0009-0000-3498-4077}{Siyu Li}$^{2,\dagger}$\\[-0.1em]
\authororcid{0009-0004-0121-9028}{Hanpei Guo}$^{3}$\hspace{0.6em}\authororcid{0009-0005-5132-6478}{Jingrong Gao}$^{4}$\hspace{0.6em}\authororcid{0000-0003-2355-288X}{Zhixu Li}$^{1,2,\dagger}$}
\newcommand{\paperaffiliations}{$^{1}$ School of Information, Renmin University of China, Beijing, China\\
$^{2}$ School of Smart Governance, Renmin University of China, Suzhou, China\\
$^{3}$ School of Art and Design, Beijing Forestry University, Beijing, China\\
$^{4}$ School of Information and Software Engineering, University of Electronic Science and Technology of China, Chengdu, China}
\newcommand{\paperemails}{\href{mailto:xilinghuiye@ruc.edu.cn}{\nolinkurl{xilinghuiye@ruc.edu.cn}}\quad \href{mailto:hy.liu25@ruc.edu.cn}{\nolinkurl{hy.liu25@ruc.edu.cn}}\\
\href{mailto:siyu.li@ruc.edu.cn}{\nolinkurl{siyu.li@ruc.edu.cn}}\quad \href{mailto:ghp241401305@bjfu.edu.cn}{\nolinkurl{ghp241401305@bjfu.edu.cn}}\\
\href{mailto:2023091601025@std.uestc.edu.cn}{\nolinkurl{2023091601025@std.uestc.edu.cn}}\quad \href{mailto:zhixuli@ruc.edu.cn}{\nolinkurl{zhixuli@ruc.edu.cn}}}
\newcommand{\papernotes}{$^{*}$: Equal contribution.\quad $^{\dagger}$: Corresponding author.}
\newcommand{\paperkeywords}{MLLM, Code generation, Data visualization}
\newcommand{\paperccs}{Computing methodologies~Artificial intelligence; Human-centered computing~Visualization}

\graphicspath{{Chartint_NIPS/}}

\newcommand{\cmark}{\ding{51}}
\newcommand{\xmark}{\ding{55}}

\AtBeginDocument{}

\AtBeginDocument{}

\hypersetup{pdftitle={Beyond Static Charts: Benchmarking Interactive Chart-to-Code Generation for MLLMs},pdfauthor={Yuhan Bian; Haoyu Liu; Siyu Li; Hanpei Guo; Jingrong Gao; Zhixu Li}}
\begin{document}
\thispagestyle{empty}

\noindent
\begin{minipage}[c]{0.55\linewidth}
  \raggedright
  \includegraphics[height=1.5cm]{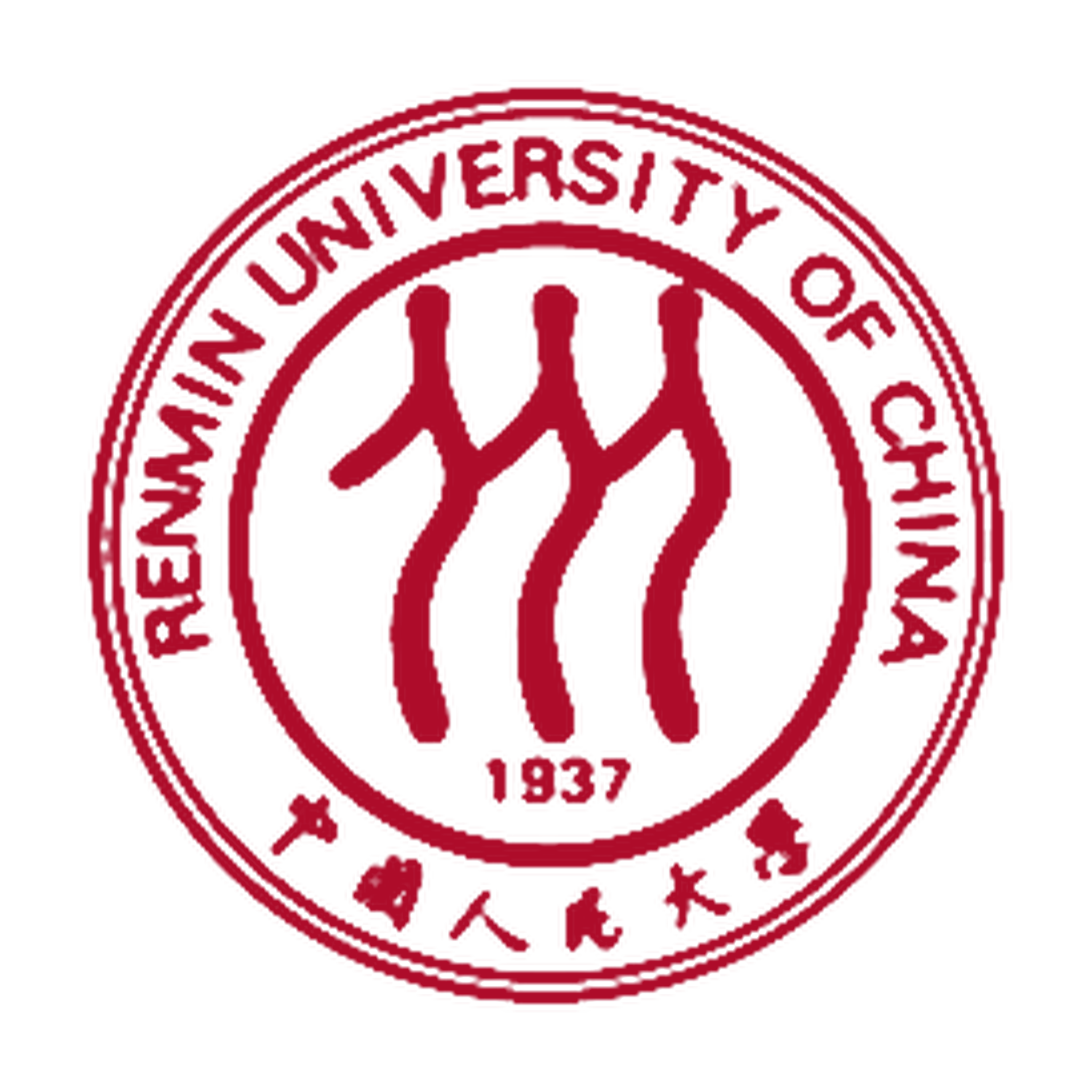}
\end{minipage}%
\begin{minipage}[c]{0.44\linewidth}
  \raggedleft
  {\rucsans\small\color{rucgray}Renmin University of China\par}

\end{minipage}\par

\vspace{0.35em}
\noindent{\color{rucline}\rule{\linewidth}{0.8pt}\par}
\vspace{1.0em}

\begin{center}
  {\ructitlefont\fontsize{13}{20}\selectfont\color{rucdark}
   \papertitle\par}
  \vspace{1.25em}
  {\normalsize\rmfamily\color{rucdark}\paperauthors\par}
  \vspace{0.22cm}
  {\footnotesize\rmfamily\color{rucgray}\paperaffiliations\par}
  {\footnotesize\rmfamily\itshape\color{rucgray}\papernotes\par}
  \vspace{0.12cm}
  {\scriptsize\rmfamily\color{rucgray}\paperemails\par}
\end{center}

\vspace{0.45em}
\begin{ructitlebox}
  \setlength{\parindent}{0cm}
  \setlength{\parskip}{0.14cm}
  \raggedright
  \nohyphens{
\label{sec:abs}
In recent years, multimodal large language models (MLLMs) have achieved significant progress in chart-to-code generation. However, existing benchmarks mainly focus on static charts, overlooking interactive charts that enable richer and more expressive data exploration.
Unlike static chart-to-code generation, which emphasizes visual consistency, interactive chart-to-code evaluation targets behavioral consistency, requiring chart states to evolve correctly under sequences of user interactions, thereby rendering traditional one-step evaluation paradigms ineffective. 
To address this gap, we introduce \textbf{ChartInt}, the first benchmark specifically designed for interactive chart-to-code generation. ChartInt consists of 2,779 carefully annotated examples covering common interaction patterns such as hover, highlight, and tooltips. Additionally, ChartInt incorporates editing tasks such as data updates, style adjustments, and cross-chart style transfer to expand real-world application scenarios.
To enable reliable evaluation of interaction behavior, we propose a preset state snapshot protocol that predefines target interaction states and deterministically triggers interaction events, transforming dynamic behaviors into reproducible evaluation targets. Based on this protocol, we design a multi-dimensional evaluation framework measuring semantic and stylistic consistency. Experiments show that while current MLLMs perform well on static charts, their performance drops significantly on interactive tasks, highlighting the need for interaction-aware benchmarks. Our code and data are available at the project repository at \url{https://github.com/bianyh/ChartInt}.
}
  \vspace{0.14cm}
  \vspace{-0.08cm}
\end{ructitlebox}

{\footnotesize\noindent\textbf{CCS Concepts:} \paperccs\par
\par}
\medskip

\section{Introduction}
\label{sec:intro}
Recent advances in multimodal large language models (MLLMs) have extended their capabilities beyond visual recognition to executable code generation\cite{chen2021evaluating, austin2021program, beltramelli2018pix2code, si2025design2code, li2024mmcode, xue2025matcha, roziere2023code, guo2024deepseek}. Chart-to-code generation aims to infer the underlying plotting logic from a chart image and produce code that reconstructs the visualization\cite{zhang2024gpt, wu2025plot2code, zhao-etal-2025-chartcoder, li-etal-2025-metal, jia2025chartreasoner, satyanarayan2015reactive}, enabling practical applications such as visualization reuse, data analysis workflows, and intelligent analytics systems\cite{savva2011revision}. To evaluate the capabilities of MLLMs in chart-to-code generation, several benchmarks have been proposed to standardize datasets and evaluation protocols\cite{wu2025plot2code, chartmimic, pesaran-zadeh-etal-2024-text2chart31, chen2024viseval}. However, existing benchmarks predominantly focus on static charts, where the evaluation objective is \textbf{visual consistency}, that is, to reproduce a visually similar figure from a given image. 

\begin{figure}[htbp]
    \centering
    \includegraphics[width=1\linewidth]{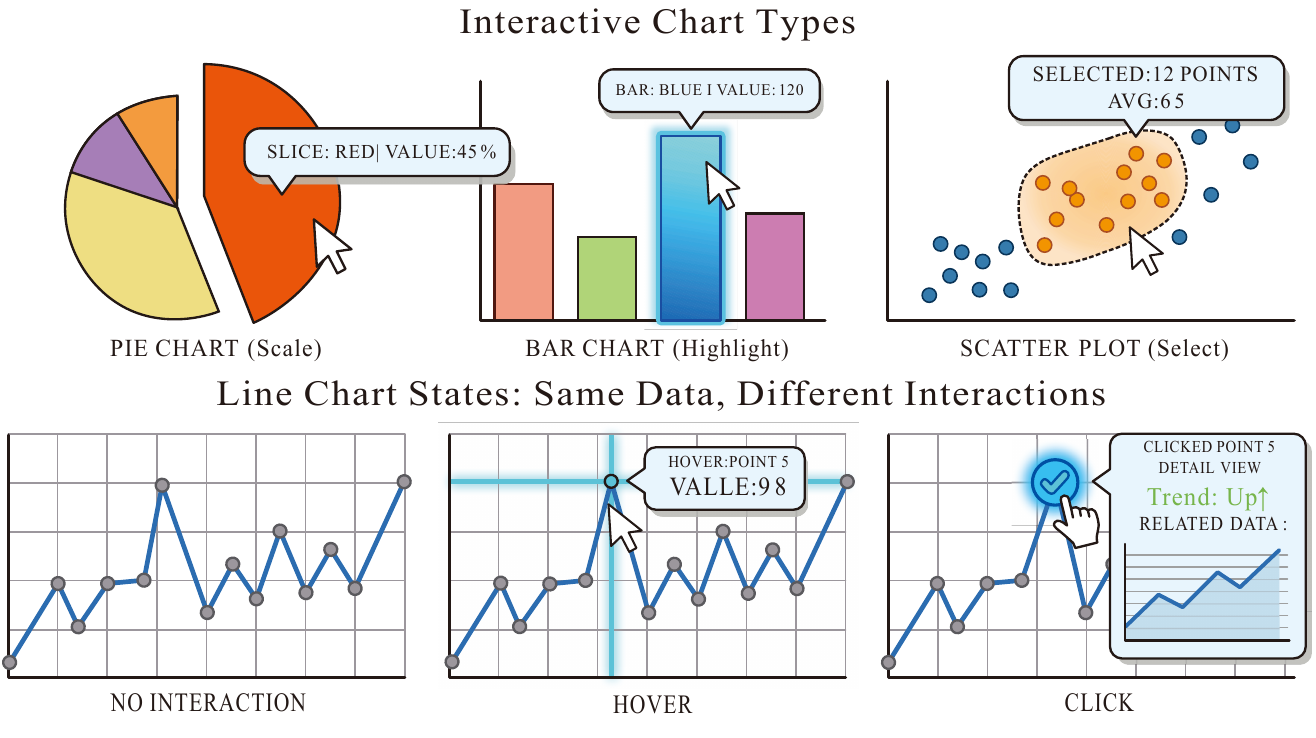}
    \caption{
    Interaction primitives and state transitions. Top: Diverse interaction types across charts. Bottom: Dynamic evolution of interaction states from static to hover and click.}
    \label{fig:py-web}
    \vspace{-2mm}
    \Description{Examples of chart interaction primitives and state transitions from static views to hover and click states.}
    \vspace{-2mm}
\end{figure}

Compared to static charts, interactive charts are characterized by executable interaction logic and dynamic state evolution, which play a critical role in real-world data exploration and analysis~\cite{heer2012interactive, li2018echarts, yi2007toward, shneiderman2003eyes}. As illustrated in Fig. \ref{fig:py-web}, interactive charts go beyond static rendering. Interactions such as hovering, clicking and selecting trigger dynamic state transitions, transforming static visualizations into explorable mediums that facilitate data analysis. Consequently, evaluating interactive chart-to-code generation requires not only examining visual consistency but also assessing event-driven behavior and state transitions across multiple interaction paths—what we refer to as \textbf{behavioral consistency}. Evaluating interactive chart-to-code generation presents two major challenges. First, interactive patterns are highly diverse and involve complex state transitions, lacking standardized evaluation data for interactive charts. Second, interactions lead to intricate dynamic state changes, making it difficult to determine whether MLLMs correctly capture the relationship between user actions and chart responses based on static outputs.

To address these challenges, we introduce \textbf{ChartInt}, the first benchmark specifically designed to evaluate interactive chart-to-code generation in MLLMs. ChartInt is built on the ECharts framework \cite{li2018echarts} and evaluates model capabilities beyond static visual reconstruction. To enable reliable assessment of interaction behaviors, we propose a preset state snapshot protocol that deterministically activates target interaction states and captures their rendered results as reproducible visual snapshots, converting dynamic interactions into stable evaluation targets. In terms of benchmark task design, ChartInt includes static reconstruction tasks for baseline comparison, as well as interaction understanding and realistic editing tasks, such as data updates, style adjustments, and cross-chart style transfer. In particular, we are the first to introduce cross-chart style transfer into chart-to-code evaluation, requiring models to preserve data semantics while transferring visual encoding rules to new chart styles. This task imposes stronger demands on both semantic understanding and visual expression. Through this multi-task framework, ChartInt comprehensively evaluates model usability in real-world web visualization generation. As shown in Table \ref{tab:benchmark_comparison}, ChartInt fills critical gaps in existing benchmarks by incorporating interaction evaluation, data updates, and style transfer. Our contributions are summarized as follows:
\begin{itemize}
    \item We are the first to highlight the importance of evaluating MLLMs on interactive chart-to-code generation, and we systematically annotated a diverse set of data samples to support this evaluation.
    
    \item To ensure chart states evolve correctly under sequences of user interactions, we propose a preset state snapshot protocol that enables deterministic triggering and reproducible evaluation of dynamic interaction behaviors.

    \item We design a suite of realistic chart editing tasks, including data updates, style modifications, and cross-chart style transfer, along with corresponding evaluation methodologies.

    \item We conduct extensive experiments on a range of state-of-the-art models, demonstrating that strong performance in static visual reconstruction does not necessarily translate into correct interaction behaviors or reliable real-world usability.
\end{itemize}

\section{Related Work}
\label{sec:rw}
\paragraph{\textbf{Multimodal Large Language Models for Chart Understanding.}}
Chart understanding tasks aim to evaluate a model’s ability to extract meaningful information from visualized data. Early benchmarks, such as FigureQA\cite{kahou2017figureqa} and DVQA\cite{kafle2018dvqa}, rely on synthetic charts paired with template-based questions, primarily focusing on basic visual reasoning. To better approximate real-world scenarios, PlotQA\cite{methani2020plotqa} and ChartBench\cite{xu2023chartbench} introduced charts constructed from real-world data sources, placing greater emphasis on numerical computation, multi-step reasoning, and complex logical inference.
With the emergence of multimodal large language models, recent works such as ChartLlama\cite{han2023chartllama} and ChartAssistant\cite{meng2024chartassisstant} leverage instruction tuning to enhance model performance across a broad range of chart-centric tasks, including chart question answering, summarization, and numerical reasoning. All these results demonstrate that MLLMs possess strong chart perception and reasoning capabilities\cite{masry2022chartqa, liu2023deplot, masry2023unichart, xia2025chartx, wang2024charxiv, yang2024matplotagent}.

\setlength{\tabcolsep}{2pt} 
\begin{table}[t]
\centering
\caption{
Comparison with existing benchmarks. 
ChartInt uniquely fills critical gaps by incorporating interaction evaluation and 3 real-world editable tasks.
}
\vspace{-2mm}
\small
\begin{adjustbox}{max width=\linewidth,max totalheight=0.80\textheight}
\begin{tabular}{lcccc}
\toprule
 &  & \multicolumn{3}{c}{\textbf{Real-world Editable}} \\ 
\cmidrule(lr){3-5} 
\textbf{Benchmark} 
& \textbf{Interaction} 
& \textbf{Soft-coding}          
& \textbf{Style}  
& \textbf{Cross-Chart}   
\\
& 
& \textbf{Data Update} 
& \textbf{Adjust} 
& \textbf{Style Transfer} \\
\midrule

ChartX\cite{xia2025chartx}
& \textcolor{red!75!black}\xmark
& \textcolor{red!75!black}\xmark
& \textcolor{red!75!black}\xmark
& \textcolor{red!75!black}\xmark \\

Plot2Code\cite{wu2025plot2code}
& \textcolor{red!75!black}\xmark
& \textcolor{red!75!black}\xmark
& \textcolor{red!75!black}\xmark
& \textcolor{red!75!black}\xmark \\

ChartMimic\cite{chartmimic}
& \textcolor{red!75!black}\xmark
& \textcolor{red!75!black}\xmark
& \textcolor{red!75!black}\xmark
& \textcolor{red!75!black}\xmark \\

Chart2Code\cite{tang2026chartscodehierarchicalbenchmark}
& \textcolor{red!75!black}\xmark
& \textcolor{green!55!black}\cmark  
& \textcolor{green!55!black}\cmark  
& \textcolor{red!75!black}\xmark  
\\

ChartM$^3$\cite{yang2025chartm3benchmarkingchartediting}
& \textcolor{red!75!black}\xmark
& \textcolor{red!75!black}\xmark  
& \textcolor{green!55!black}\cmark  
& \textcolor{red!75!black}\xmark  
\\

\midrule
\textbf{ChartInt (Ours)} 
& \textcolor{green!55!black}\cmark
& \textcolor{green!55!black}\cmark
& \textcolor{green!55!black}\cmark
& \textcolor{green!55!black}\cmark \\
\bottomrule
\end{tabular}
\end{adjustbox}
\label{tab:benchmark_comparison}
\end{table}

\paragraph{\textbf{Chart-to-Code Benchmarks.}}
Unlike passive chart understanding, chart-to-code generation requires models to synthesize executable charting code directly from visual inputs, constituting a substantially more challenging inverse graphics task. To standardize evaluation in this setting, Plot2Code\cite{wu2025plot2code} introduces code executability as a prerequisite constraint and performs dual-view evaluation by combining text-level assessment via PaddleOCR\cite{du2020pp} with visual comparison using GPT-4V\cite{achiam2023gpt}. 
ChartMimic\cite{chartmimic} further develops a fine-grained evaluation framework that includes low-level metrics over textual content, layout, chart type, and color, together with high-level holistic scoring conducted by GPT-4o, with a primary focus on visual fidelity in static chart reproduction.
Despite these advances, existing chart-to-code benchmarks largely emphasize static visual consistency. This leaves a gap for executable chart code whose correctness depends on whether specified interaction states can be reached after execution.

\section{Problem Formulation}
ChartInt formulates chart reproduction and editing as conditional program synthesis, requiring executable code to preserve visual semantics and interactive behavior. Editing additionally specifies intended changes and invariants. Correctness depends on rendered outcomes rather than source-code identity.

\begin{figure}[t]
    \centering
    \includegraphics[width=1\linewidth]{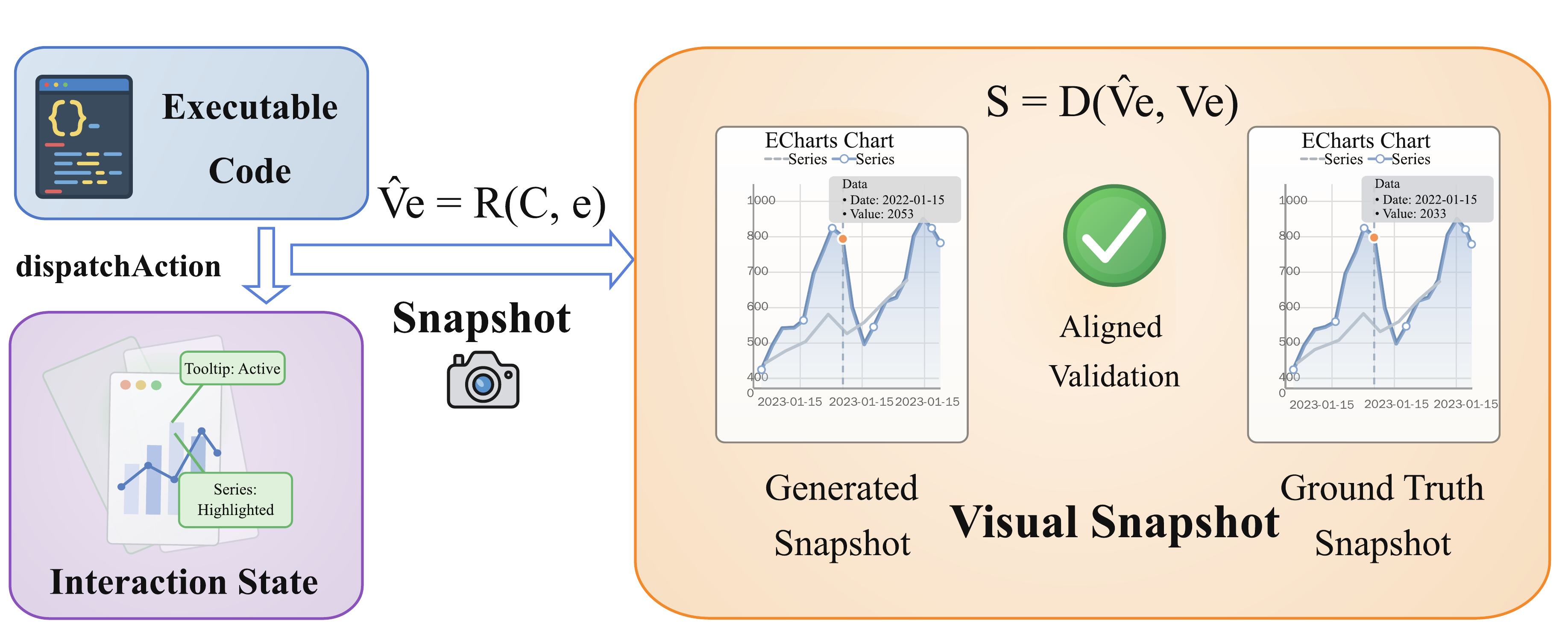}
    \caption{Preset state snapshot protocol. Programmatic API calls deterministically activate predefined interaction states, which are captured as rendered snapshots for reproducible evaluation.}
    \label{fig:pssp}
\end{figure}

\subsection{Interactive Chart Reproduction}

Given a chart image $V$, a model $f_{\mathrm{MLLM}}$ generates executable code $C$. The rendering function $\mathcal{R}$ produces the chart's visual state, and $\mathcal{D}$ measures its discrepancy from the reference:

\begin{equation}
C = f_{\mathrm{MLLM}}(V);\quad S(C, V) = \mathcal{D}\Big( \mathcal{R}(C), \; V \Big).
\end{equation}
Behavioral consistency further conditions rendering on interaction events, as formalized in Section~\ref{sec:pssp}.

\subsection{Interactive Chart Editing}

Editing conditions generation on the original image $V$ and instruction $T$. The modified program $C'$ must render the intended target $V'$ while preserving unedited content and behavior:

\begin{equation}
C' = f_{\mathrm{MLLM}}(V, T); \quad S(C', V') = \mathcal{D}\Big( \mathcal{R}(C'), \; V' \Big).
\end{equation}

\section{Dataset Construction and Evaluation}

To ensure a realistic and comprehensive evaluation, ChartInt is built upon carefully curated datasets, diverse interaction types, and a unified task taxonomy. In this section, we present the dataset construction process and detail the specific benchmark tasks: static reconstruction, interaction verification, and editable modification.

\begin{table}[t]
\centering
\small
\caption{Task distribution of ChartInt. The benchmark contains 2,779
instances across six tasks, covering static reconstruction, interaction
verification, and real-world chart editing.}
\label{tab:chartint_statistics}
\begin{adjustbox}{max width=\linewidth,max totalheight=0.80\textheight}
\begin{tabular}{@{}llr@{}}
\toprule
\textbf{Task Category} & \textbf{Task} & \textbf{\#Samples} \\
\midrule
\multirow{2}{*}{Static Reconstruction}
& Native Reconstruction & 557 \\
& Perturbed Reconstruction & 537 \\
\midrule
\textbf{Interaction}
& Interaction & 375 \\
\midrule
\multirow{3}{*}{Real-World Editable}
& Data Update & 431 \\
& Style Adjustment & 451 \\
& Style Transfer & 428 \\
\midrule
\multicolumn{2}{r}{\textbf{Total}}
& \textbf{2,779} \\
\bottomrule
\end{tabular}
\end{adjustbox}
\end{table}

\subsection{Data Pipeline \& Static Reconstruction Tasks}\label{LLM-driven Data Augmentation and Synthesis} 
\begin{figure*}[t]
    \centering
    \includegraphics[width=1\linewidth]{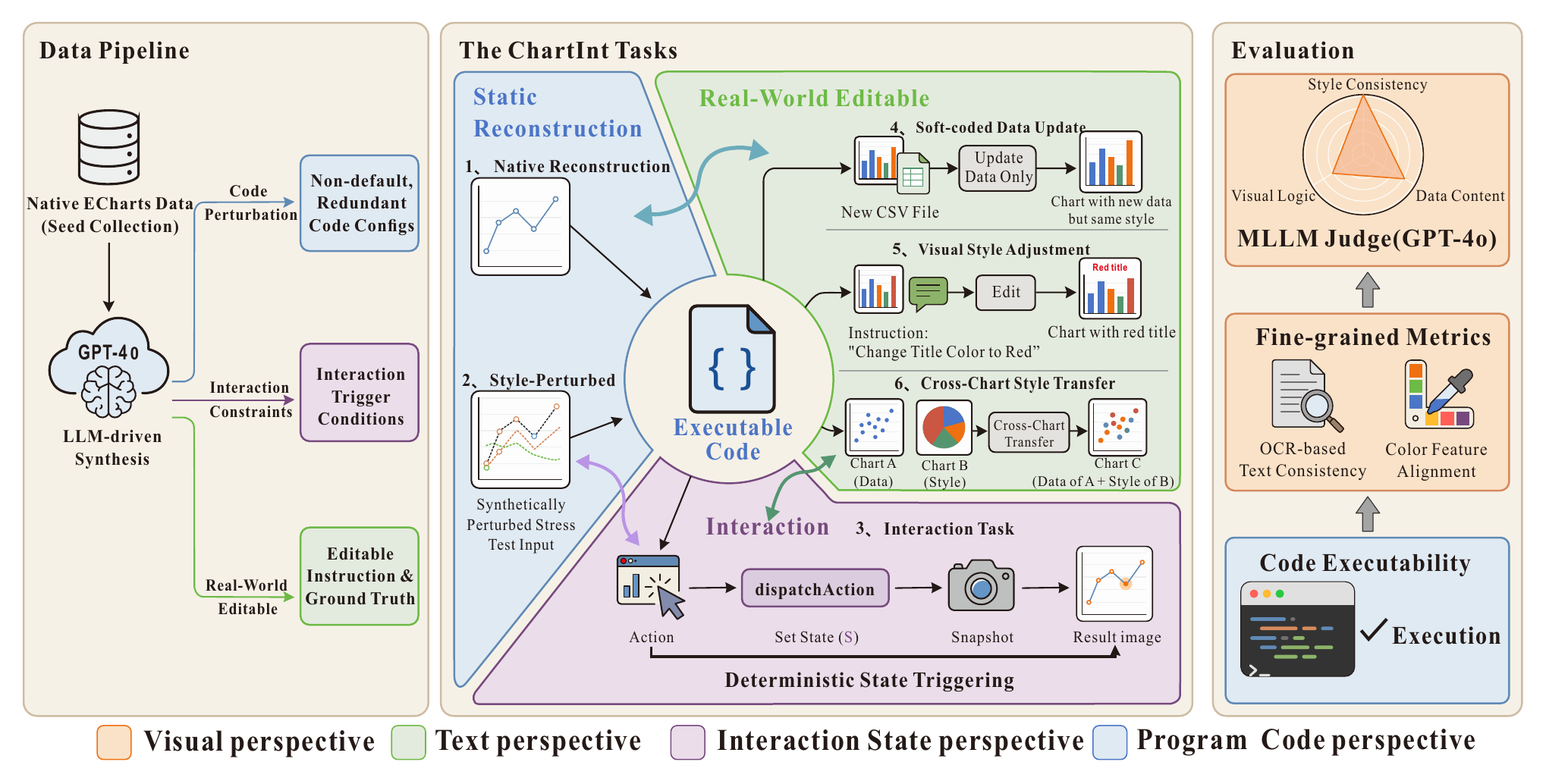}
    \caption{Overview of the ChartInt framework. The architecture utilizes an LLM-driven Data Pipeline to synthesize benchmark assets. The central ChartInt Tasks cover three categories: Static Reconstruction (Native \& Perturbed), Interaction (verifying dynamic behaviors via deterministic state triggering), and Real-World Editable (Data Update, Style Adjustment, and Style Transfer). Finally, the Evaluation Suite assesses performance using Code Executability, Fine-grained Metrics (OCR, Color), and MLLM-based judgment.}
    \label{fig:architecture}
    \Description{Pipeline diagram showing data synthesis, ChartInt task categories, and the evaluation suite.}
\end{figure*}

We construct ChartInt by curating high-quality ECharts from the community, filtering for rendering consistency to form a seed dataset. Building on this foundation, we employ GPT-4o for program-level synthesis\cite{huang2024evochartbenchmarkselftrainingapproach} to generate evaluation tasks spanning static reconstruction, interaction correctness, and editing (Figure~\ref{fig:architecture}). 

To contextualize interactive performance, we establish two static baselines: (1) \textbf{Native Reconstruction}: Reproducing original seed charts to assess fundamental visual-to-code mapping. (2) \textbf{Perturbed Reconstruction}: Reproducing charts with injected non-default styles (e.g., custom gradients) to stress-test robustness and identify static bottlenecks before interaction evaluation.

Table~\ref{tab:chartint_statistics} reports 2,779 task instances across six complementary tasks: static baselines, event-conditioned interaction, and three editing scenarios. These task-level results distinguish visual reconstruction, interaction logic, data manipulation, and compositional style reasoning.

\subsection{Preset State Snapshot Protocol}
\label{sec:pssp}

A fundamental challenge in evaluating interactive charts is the transient and non-deterministic nature of interaction states (e.g., hovering, selection). Manual evaluation is irreproducible, while static snapshots obtained from direct rendering are inherently incapable of reflecting dynamic behaviors.

To address this, we propose the \textbf{Preset State Snapshot Protocol}, a deterministic evaluation mechanism that maps dynamic interaction logic to observable static outputs. As illustrated in Figure \ref{fig:pssp}, instead of relying on manual triggers, this protocol programmatically invokes the visualization engine's API (dispatchAction in ECharts) to force the generated chart into a specified interaction state.

Formally, we extend the rendering function $\mathcal{R}$ to accept an Interaction event setting $e$. The evaluation process transforms the code $C$ into a state-specific snapshot $\hat{V}_e$:
\begin{equation}
\hat{V}_e = \mathcal{R}(C, e),
\end{equation}
where $e$ represents the deterministic event setting 
(e.g., \texttt{showTip} with \texttt{seriesIndex: 0} and \texttt{dataIndex: 0} targets the first data item in the first series). 

ECharts instantiates $e$ through \texttt{dispatchAction}, supporting highlighting, tooltips, legend selection, data zooming, and component-specific selection. Table~\ref{tab:echarts_actions} lists representative API actions, rather than instance counts by interaction type. Programmatic activation removes the need for manual mouse or keyboard operations. The resulting snapshot is compared with its ground-truth state using the visual validation pipeline shared with static reconstruction\cite{liu2026omniverifier}.

This protocol ensures that if the generated code $C$ correctly implements the interaction logic, the rendered snapshot $\hat{V}_e$ will visually match the ground-truth state $V_e$. This reduces the complex dynamic evaluation problem to a robust static image comparison:$ \mathcal{D}(\hat{V}_e, V_e) $.

\begin{table*}[t]
\centering
\small
\caption{Representative ECharts \texttt{dispatchAction} types supported by the
interaction evaluation protocol. The table lists common action types that can
be programmatically triggered to construct deterministic interaction states.}
\label{tab:echarts_actions}
\begin{adjustbox}{max width=\linewidth,max totalheight=0.80\textheight}
\begin{tabular}{@{}lll@{}}
\toprule
\textbf{Category} & \textbf{Action Type} & \textbf{Description} \\
\midrule
\multirow{2}{*}{\textbf{Highlight}}
& \texttt{highlight}
& Highlights specified data series or items. \\
& \texttt{downplay}
& Cancels the highlight state. \\
\midrule

\multirow{2}{*}{\textbf{Tooltip}}
& \texttt{showTip}
& Manually triggers a tooltip at a specified location or data item. \\
& \texttt{hideTip}
& Hides the currently displayed tooltip. \\
\midrule

\multirow{2}{*}{\textbf{Legend}}
& \texttt{legendSelect}
& Selects a legend item and displays the corresponding series. \\
& \texttt{legendToggleSelect}
& Toggles the selection state of a legend item. \\
\midrule

\multirow{2}{*}{\textbf{DataZoom}}
& \texttt{dataZoom}
& Zooms into a specified range of the coordinate system. \\
& \texttt{takeGlobalCursor}
& Activates the brush or zoom cursor tool. \\
\midrule

\multirow{2}{*}{\textbf{Pie/Sunburst}}
& \texttt{select}
& Selects a specified data item, e.g., a pie or sunburst segment. \\
& \texttt{unselect}
& Deselects a previously selected data item. \\
\midrule

\multirow{2}{*}{\textbf{Map}}
& \texttt{geoSelect}
& Selects a specific region in a geographic coordinate system. \\
& \texttt{geoRoam}
& Controls panning and zooming of the map. \\
\midrule

\textbf{VisualMap}
& \texttt{selectDataRange}
& Adjusts the selected range of the visualMap component. \\
\bottomrule
\end{tabular}
\end{adjustbox}
\end{table*}

\paragraph{State paths and semantic targets.}
At the protocol level, for a path $\boldsymbol{e}=(e_1,\ldots,e_m)$, the same contract applies actions in order and captures the resulting state; an independent preset is the case $m=1$. Thus the protocol represents both individual target states and ordered transitions. A visible preset state and the user gesture that reaches it are distinct: API activation specifies the state to compare, while the generated program supplies its interaction logic. Event indices identify runtime components, not semantic identities. To compare reordered programs, an adapter must resolve the intended series and data item by stable identifiers, names, or data keys before dispatch; a positional index alone cannot establish semantic alignment.

\paragraph{Adapting PSSP beyond ECharts.}
The reusable contract is to identify the semantic target, activate its state, wait for rendering, and capture the same visual region in reference and generated programs. ECharts supplies the \texttt{dispatchAction} adapter. Conceptually, a Vega-Lite adapter would compile the specification to Vega, map the target parameter or selection to the compiled runtime signals or stores, apply its state update, await \texttt{view.runAsync()}, and capture the page, including any tooltip overlay~\cite{vegaViewAPI}. Plotly exposes programmatic hover and layout updates; D3 provides application-specific event handlers and brush/zoom controls~\cite{plotlyHoverAPI,plotlyFunctionsAPI,d3DispatchAPI,d3BrushAPI,d3ZoomAPI}. These are adaptation routes, not additional evaluated implementations; the state-conditioned comparison remains unchanged.

\subsection{Interaction}

The interaction task conditions generation on a chart image $V$ and a description $T_{\text{state}}$ of the required interaction state:

\begin{equation}
C_{\text{interaction}} = f_{\mathrm{MLLM}}(V, T_{\text{state}}).
\end{equation}
For the corresponding deterministic event $e$, interaction correctness compares the event-conditioned rendering with the reference state $V_e$:

\begin{equation}
S(C_{\text{interaction}}, V_e) = 
\mathcal{D}\Big( \mathcal{R}(C_{\text{interaction}}, e_{\text{programmatically}}), \; V_e \Big).
\end{equation}
This tests whether the generated program reaches the requested interaction state, using reproducible visual consistency checks.

\subsection{Real-World Editable Tasks}

Beyond reproduction, ChartInt further evaluates a model’s ability to modify charts under explicit constraints. We formulate chart editing as a constrained generation problem where the goal is to produce modified code $\hat{C}'$ that satisfies a specific transformation $\mathcal{T}$ while preserving the invariant components of the original chart. Table~\ref{tab:editing_contracts} makes these constraints explicit.

\begin{table}[t]
\centering
\small
\caption{Requested changes and invariants in editing tasks.}
\label{tab:editing_contracts}
\begin{adjustbox}{max width=\linewidth,max totalheight=0.80\textheight}
\begin{tabular}{@{}p{0.17\columnwidth}p{0.35\columnwidth}p{0.40\columnwidth}@{}}
\toprule
\textbf{Task} & \textbf{Change} & \textbf{Preserve} \\
\midrule
Data update & Load the supplied external data. & Chart type, data-to-channel mapping, and visual style. \\
Style \newline adjustment & Apply the instructed visual attributes. & Data values, field meanings, and chart structure. \\
Style \newline transfer & Apply compatible appearance attributes from $V_B$. & Data semantics, chart type, axis semantics, and mark type of $V_A$. \\
\bottomrule
\end{tabular}
\end{adjustbox}
\end{table}

\subsubsection{Data Update.}

Given an original chart image $V$ and a textual instruction $T_{\text{data}}$ specifying the target data source, the data update task requires the model to modify the underlying data loading logic of the chart. A key constraint of this task is soft-coding: the generated program must dynamically load external data files (e.g., via file I/O APIs) rather than hard-coding numerical values directly into the code.

In this setting, the updated data is provided as an external file set $\mathcal{F}_{\text{csv}}$ within the execution environment. Based on the input image and instruction, the model generates executable chart code $C_{\text{data}}$, which is evaluated by executing the generated program and comparing the rendered result against the ground-truth $V'$:
\begin{equation}
C' = f_{\mathrm{MLLM}}(V, T_{\text{data}}); \quad
S(C', V') = \mathcal{D}\Big( \mathcal{R}(C', \mathcal{F}_{\text{csv}}), \; V' \Big).
\end{equation}
Here, $\mathcal{R}(C, \mathcal{F}_{\text{csv}})$ denotes executing code $C$ in a runtime environment where the external file set $\mathcal{F}_{\text{csv}}$ is available. The discrepancy function $\mathcal{D}$ evaluates the correctness and visual consistency of the updated data. Soft-coding remains a program-level requirement: matching the rendered target alone does not establish that the program loads external data.

\subsubsection{Style Adjustment.}

This task focuses on modifying the aesthetic appearance of a chart without altering its underlying data or structural logic. Supported edits include changes to color schemes, font styles, axis attributes, layout coordinates, and legend positioning.

Let $T_{\text{style}}$ denote a textual instruction specifying the desired visual changes (e.g., ``Change the bar color to red''). Conditioned on the input chart image and instruction, the model generates executable chart code $C_{\text{style}}$. Correctness is assessed under an edit locality constraint by comparing the rendered code against a target state $V'_{\text{style}}$ which reflects the new style applied to the original data:
\begin{equation}
C' = f_{\mathrm{MLLM}}(V, T_{\text{style}}); \quad
S(C', V') = \mathcal{D}\Big( \mathcal{R}(C'), \; V' \Big).
\end{equation}

\subsubsection{Style Transfer.}

Cross-chart style transfer evaluates the model's ability to perform compositional reasoning across multiple visualization contexts. Let $V_A$ denote the content source image (specifying data semantics) and $V_B$ denote the style reference image (defining visual appearance). The task requires generating a new chart that combines the content of $V_A$ with the style of $V_B$. Chart type, mark type, and the axes' semantic roles belong to the content source; colors, typography, and compatible presentation attributes belong to style. Axis styling and positioning may change, but replacing the source's data-to-axis relationships or converting a line into bars is not style transfer. When inputs differ structurally, the reference supplies compatible appearance cues rather than a replacement data representation.

We formulate this as a dual-input generation problem where the model produces code $C_{\text{transfer}}$. To assess whether content and style are correctly disentangled and recombined, we execute the generated program and compare the result against a ground-truth composition $V_{A \rightarrow B}$:
\begin{equation}
C' = f_{\mathrm{MLLM}}(V_A, V_B); \quad
S(C', V_{A \rightarrow B}) = \mathcal{D}\Big( \mathcal{R}(C'), \; V_{A \rightarrow B} \Big).
\end{equation}

\definecolor{top1_gold}{HTML}{FFD700}
\definecolor{top2_silver}{HTML}{DCDCDC}
\definecolor{top3_bronze}{HTML}{D2B48C}

\newcommand{\first}[1]{\colorbox{top1_gold}{\textbf{#1}}}
\newcommand{\second}[1]{\colorbox{top2_silver}{\textbf{#1}}}
\newcommand{\third}[1]{\colorbox{top3_bronze}{\textbf{#1}}}

\begin{table*}[t] 
\centering 
\caption{Fine-grained evaluation results across the six ChartInt tasks. We report objective visual metrics (Text Consistency and Color Alignment) and GPT-based perceptual scores (Style, Visual Logic, and Data Faithfulness). For each column, \first{first}, \second{second}, and \third{third} ranked models are highlighted with gold, silver, and bronze background colors, respectively.} 
\resizebox{\textwidth}{!}{ 
\begin{adjustbox}{max width=\linewidth,max totalheight=0.80\textheight}
\begin{tabular}{l|ccccc|ccccc|ccccc|ccccc|ccccc|ccccc} 
\toprule 
\multirow{2}{*}{\textbf{Model}} & \multicolumn{5}{c}{\textbf{Native Reconstruction}} & \multicolumn{5}{c}{\textbf{Perturbed Reconstruction}} & \multicolumn{5}{c}{\textbf{Interaction}} & \multicolumn{5}{c}{\textbf{Data Update}} & \multicolumn{5}{c}{\textbf{Style Adjustment}} & \multicolumn{5}{c}{\textbf{Style Transfer}} \\ 
\cmidrule(lr){2-6} \cmidrule(lr){7-11} \cmidrule(lr){12-16} \cmidrule(lr){17-21} \cmidrule(lr){22-26} \cmidrule(lr){27-31} 
& \multicolumn{2}{c}{\textbf{Detail}} & \multicolumn{3}{c}{\textbf{GPT}} & \multicolumn{2}{c}{\textbf{Detail}} & \multicolumn{3}{c}{\textbf{GPT}} & \multicolumn{2}{c}{\textbf{Detail}} & \multicolumn{3}{c}{\textbf{GPT}} & \multicolumn{2}{c}{\textbf{Detail}} & \multicolumn{3}{c}{\textbf{GPT}} & \multicolumn{2}{c}{\textbf{Detail}} & \multicolumn{3}{c}{\textbf{GPT}} & \multicolumn{2}{c}{\textbf{Detail}} & \multicolumn{3}{c}{\textbf{GPT}} \\ 
\cmidrule(lr){2-3} \cmidrule(lr){4-6} \cmidrule(lr){7-8} \cmidrule(lr){9-11} \cmidrule(lr){12-13} \cmidrule(lr){14-16} \cmidrule(lr){17-18} \cmidrule(lr){19-21} \cmidrule(lr){22-23} \cmidrule(lr){24-26} \cmidrule(lr){27-28} \cmidrule(lr){29-31} 
& Text & Color & Sty & Vis & Data & Text & Color & Sty & Vis & Data & Text & Color & Sty & Vis & Data & Text & Color & Sty & Vis & Data & Text & Color & Sty & Vis & Data & Text & Color & Sty & Vis & Data \\ 
\midrule 
Qwen3-VL-2B & 0.452 & 0.284 & 2.352 & 1.935 & 2.407 & 0.506 & 0.257 & 2.119 & 2.095 & 2.548 & 0.477 & 0.295 & 2.293 & 1.767 & 1.977 & 0.318 & 0.265 & 2.185 & 1.796 & 1.796 & 0.428 & 0.282 & 2.156 & 1.701 & 1.610 & 0.114 & 0.160 & 2.000 & 1.000 & 1.000 \\
Qwen3-VL-4B & 0.564 & 0.347 & 2.586 & 2.623 & 3.152 & 0.393 & 0.279 & 2.332 & 2.652 & 3.275 & 0.497 & 0.357 & 2.570 & 2.544 & 2.620 & 0.415 & 0.333 & 2.356 & 2.175 & 2.296 & 0.583 & 0.348 & 2.604 & 2.469 & 2.798 & 0.207 & 0.286 & 2.165 & 1.504 & 1.424 \\
Qwen3-VL-8B & 0.605 & 0.363 & 2.626 & 2.428 & 3.114 & 0.547 & 0.308 & 2.507 & 2.362 & 2.918 & 0.504 & 0.357 & 2.597 & 2.597 & 2.626 & 0.493 & 0.343 & 2.527 & 2.070 & 2.411 & 0.588 & 0.343 & 2.526 & 2.285 & 2.751 & 0.211 & 0.285 & 2.149 & 1.497 & 1.427 \\
Qwen3-VL-32B & 0.724 & 0.390 & 3.071 & 3.024 & 3.654 & 0.660 & 0.349 & 3.017 & 3.160 & 3.567 & 0.603 & 0.399 & 3.015 & 3.104 & 2.675 & 0.520 & 0.374 & 2.647 & 2.279 & 2.324 & 0.630 & 0.376 & 2.792 & 2.744 & 2.980 & \third{0.562} & 0.328 & 2.431 & 1.976 & \third{2.103} \\
Qwen3-VL-235B & 0.749 & 0.405 & 3.040 & 3.149 & 3.733 & 0.683 & 0.346 & 3.128 & 3.365 & 3.703 & 0.679 & \third{0.439} & 3.120 & 3.120 & 2.913 & \third{0.641} & 0.416 & 2.833 & 2.655 & 2.774 & 0.695 & 0.379 & 2.853 & 2.853 & 3.211 & 0.432 & \third{0.341} & 2.354 & 1.823 & 1.938 \\
\midrule 
InternVL3.5-2B & 0.522 & 0.287 & 2.173 & 1.945 & 2.332 & 0.462 & 0.250 & 2.006 & 1.892 & 2.375 & 0.472 & 0.291 & 2.103 & 1.877 & 2.256 & 0.295 & 0.275 & 2.105 & 1.544 & 1.632 & 0.439 & 0.291 & 2.095 & 1.897 & 2.011 & 0.336 & 0.290 & 2.026 & 1.250 & 1.276 \\
InternVL3.5-8B & 0.567 & 0.284 & 2.348 & 2.179 & 2.721 & 0.532 & 0.237 & 2.091 & 2.269 & 3.020 & 0.532 & 0.293 & 2.335 & 2.238 & 2.707 & 0.431 & 0.289 & 2.251 & 2.011 & 2.149 & 0.534 & 0.294 & 2.286 & 2.159 & 2.469 & 0.184 & 0.279 & 2.059 & 1.267 & 1.211 \\
InternVL3.5-38B & 0.563 & 0.268 & 2.299 & 2.369 & 2.823 & 0.527 & 0.234 & 2.108 & 2.512 & 3.163 & 0.524 & 0.303 & 2.334 & 2.363 & 2.306 & 0.453 & 0.295 & 2.294 & 2.115 & 2.218 & 0.557 & 0.303 & 2.343 & 2.333 & 2.649 & 0.283 & 0.252 & 2.063 & 1.383 & 1.357 \\
\midrule 
Idefics2-8B & 0.000 & 0.246 & 2.009 & 1.522 & 1.478 & 0.000 & 0.268 & 1.978 & 1.593 & 1.615 & 0.000 & 0.000 & 1.000 & 1.000 & 1.000 & 0.000 & 0.000 & 1.000 & 1.000 & 1.000 & 0.000 & 0.000 & 1.000 & 1.000 & 1.000 & 0.000 & 0.269 & 1.848 & 1.087 & 1.087 \\
DeepSeek-7B & 0.000 & 0.283 & 2.000 & 1.000 & 1.000 & 0.000 & 0.364 & 2.000 & 1.000 & 1.000 & 0.000 & 0.000 & 1.000 & 1.000 & 1.000 & 0.000 & 0.000 & 1.000 & 1.000 & 1.000 & 0.000 & 0.200 & 0.000 & 0.000 & 0.000 & 0.000 & 0.000 & 1.000 & 1.000 & 1.000 \\
minicpm(~8B) & 0.398 & 0.285 & 2.066 & 1.305 & 1.427 & 0.373 & 0.252 & 1.987 & 1.283 & 1.465 & 0.378 & 0.301 & 2.037 & 1.293 & 1.366 & 0.334 & 0.295 & 2.056 & 1.319 & 1.542 & 0.323 & 0.278 & 2.015 & 1.244 & 1.267 & 0.203 & 0.268 & 2.000 & 1.143 & 1.143 \\
Phi-3(4B) & 0.004 & 0.217 & 1.929 & 1.500 & 1.214 & 0.003 & 0.242 & 1.923 & 1.615 & 2.077 & 0.000 & 0.289 & 1.800 & 1.200 & 1.467 & 0.000 & 0.000 & 1.000 & 1.000 & 1.000 & 0.000 & 0.248 & 2.000 & 1.667 & 1.889 & 0.015 & 0.268 & 1.833 & 1.000 & 1.500 \\ 
llava-7b & 0.091 & 0.272 & 2.000 & 1.033 & 1.011 & 0.091 & 0.249 & 1.986 & 1.108 & 1.027 & 0.100 & 0.268 & 2.000 & 1.061 & 1.000 & 0.148 & 0.265 & 2.023 & 1.136 & 1.159 & 0.067 & 0.261 & 2.000 & 1.035 & 1.088 & 0.090 & 0.271 & 2.000 & 1.035 & 1.000 \\
\midrule 

GPT-4o & 0.664 & 0.387 & 2.886 & 2.985 & 3.270 & 0.601 & 0.317 & 2.670 & 2.930 & 3.130 & 0.625 & 0.389 & 2.853 & 2.965 & 2.240 & 0.589 & 0.375 & 2.723 & 2.748 & 2.655 & 0.615 & 0.363 & 2.715 & 2.854 & 2.718 & 0.518 & 0.330 & 2.409 & 2.125 & 1.925 \\
claude-opus-4-6 & \second{0.835} & \third{0.442} & \third{3.820} & \third{4.500} & \third{4.440} & \third{0.813} & \third{0.367} & \third{3.787} & \third{4.468} & \second{4.617} & \third{0.730} & 0.420 & \third{3.500} & \third{4.000} & \third{3.200} & 0.634 & \third{0.431} & \third{3.087} & \third{2.913} & \third{2.957} & \second{0.740} & \third{0.396} & \third{3.571} & \third{4.306} & \second{4.082} & \second{0.580} & 0.331 & \third{2.696} & \second{2.500} & \second{2.217} \\
gpt-5.4 & \third{0.823} & \first{0.542} & \first{4.041} & \second{4.575} & \first{4.769} & \second{0.820} & \first{0.487} & \first{4.015} & \second{4.514} & \first{4.713} & \second{0.748} & \first{0.521} & \second{3.758} & \second{4.351} & \first{3.769} & \second{0.671} & \second{0.512} & \second{3.317} & \second{3.294} & \second{3.083} & \first{0.750} & \first{0.477} & \first{3.692} & \second{4.308} & \first{4.213} & 0.495 & \first{0.455} & \second{2.835} & \third{2.291} & 1.833 \\
gemini-3-pro & \first{0.865} & \second{0.538} & \second{3.948} & \first{4.678} & \second{4.690} & \first{0.825} & \second{0.439} & \second{3.898} & \first{4.667} & \third{4.510} & \first{0.795} & \second{0.510} & \first{3.881} & \first{4.687} & \second{3.567} & \first{0.682} & \first{0.518} & \first{3.397} & \first{3.687} & \first{3.427} & \third{0.732} & \second{0.458} & \second{3.651} & \first{4.390} & \third{3.993} & \first{0.584} & \second{0.405} & \first{3.024} & \first{3.184} & \first{2.368} \\

\bottomrule 
\end{tabular}
\end{adjustbox} 
} 
\label{tab:finegrained} 
\end{table*}

\subsection{Evaluation Methodology}

Prior benchmarks combine execution success, fine-grained metrics, and LLM-based judgment~\cite{chartmimic,wu2025plot2code}. ChartInt extends rendered-output evaluation to controlled interaction states, allowing semantically equivalent programs to be assessed by their observable behavior. We use three complementary perspectives:

\textbf{Code Executability.} We first verify whether the generated code executes successfully in a sandboxed environment\cite{memon2002gui, talakola2024automated}. This step filters out syntax errors and runtime exceptions, establishing a necessary lower bound on validity.

\textbf{MLLM-based Perceptual Evaluation.} We employ GPT-4o~\cite{achiam2023gpt} to assess high-level visual and semantic alignment~\cite{zheng2023judging}. The evaluator assigns 1--5 Likert scores across three dimensions: Visual Logic, Data Content, and Style Consistency. Detailed prompts and grading examples are included in the supplementary material and project repository.

\textbf{Fine-grained Visual Alignment Metrics.}
To complement perceptual judgments, we compute two specific metrics directly on the rendered visual outputs. Importantly, ChartInt does not use CLIP or MSE as the basis for interaction scoring: global image similarity alone can underweight small but semantically decisive states such as tooltip values or highlighted marks. Instead, PSSP first activates the target state, and the metrics below are computed on the resulting rendered chart.

OCR-based Text Consistency ($F_1$).
Textual fidelity (e.g., axis titles, legends, and data labels) is a cornerstone of chart accuracy. We utilize PaddleOCR\cite{du2020pp} to extract text sequences from both the reference image ($T_{ref}$) and the generated image ($T_{gen}$). To accommodate minor OCR recognition noise, we construct a similarity matrix using the Gestalt Pattern Matching algorithm. Let $s(i, j) \in [0, 1]$ denote the similarity between the $i$-th reference string and the $j$-th generated string. We determine the optimal alignment by solving the maximum weight bipartite matching problem using the Kuhn-Munkres (Hungarian) algorithm\cite{kuhn1955hungarian}:
\begin{equation}
    \text{Maximize}  \sum_{(i,j) \in \mathcal{M}} s(i, j)
\end{equation}
where $\mathcal{M}$ is the maximum-weight assignment. Only pairs in $\mathcal{M}_{0.85}=\{(i,j)\in\mathcal{M}:s(i,j)\geq0.85\}$ count as successful matches. Precision and Recall are $|\mathcal{M}_{0.85}|/|T_{gen}|$ and $|\mathcal{M}_{0.85}|/|T_{ref}|$, respectively; we report their harmonic mean as $F_1$.

Color Feature Alignment (Color-IoU).
To evaluate the stylistic fidelity of the chart’s visual encoding, we measure the overlap of the dominant color palette. We map the extracted RGB values from both images into a standardized categorical color space\cite{chu2025makes, 10.1117/12.477378}. Let $C_{ref}$ and $C_{gen}$ represent the sets of uniquely identified color categories in the reference and generated charts, respectively. The alignment is quantified using the Jaccard Index (Color-IoU\cite{munkres1957algorithms}):
\begin{equation}
    \text{Color-IoU} = \frac{|C_{ref} \cap C_{gen}|}{|C_{ref} \cup C_{gen}|}
\end{equation}
This metric emphasizes the presence of the correct thematic palette while remaining invariant to minor variations in pixel-level coordinates or rendering anti-aliasing.

These metrics are complementary: OCR measures visible textual agreement, while Color-IoU measures palette overlap. Neither alone verifies geometric layout, data-to-mark binding, or interaction semantics. PSSP exposes the requested state for comparison, and the perceptual dimensions assess its visual organization, displayed data, and style. Thus, execution success and both low- and high-level scores should be read together.

\section{Experiments}
\label{sec:exp}

\begin{figure*}[t]
    \centering
    \begin{minipage}{0.49\linewidth}
        \centering
        \includegraphics[width=\linewidth]{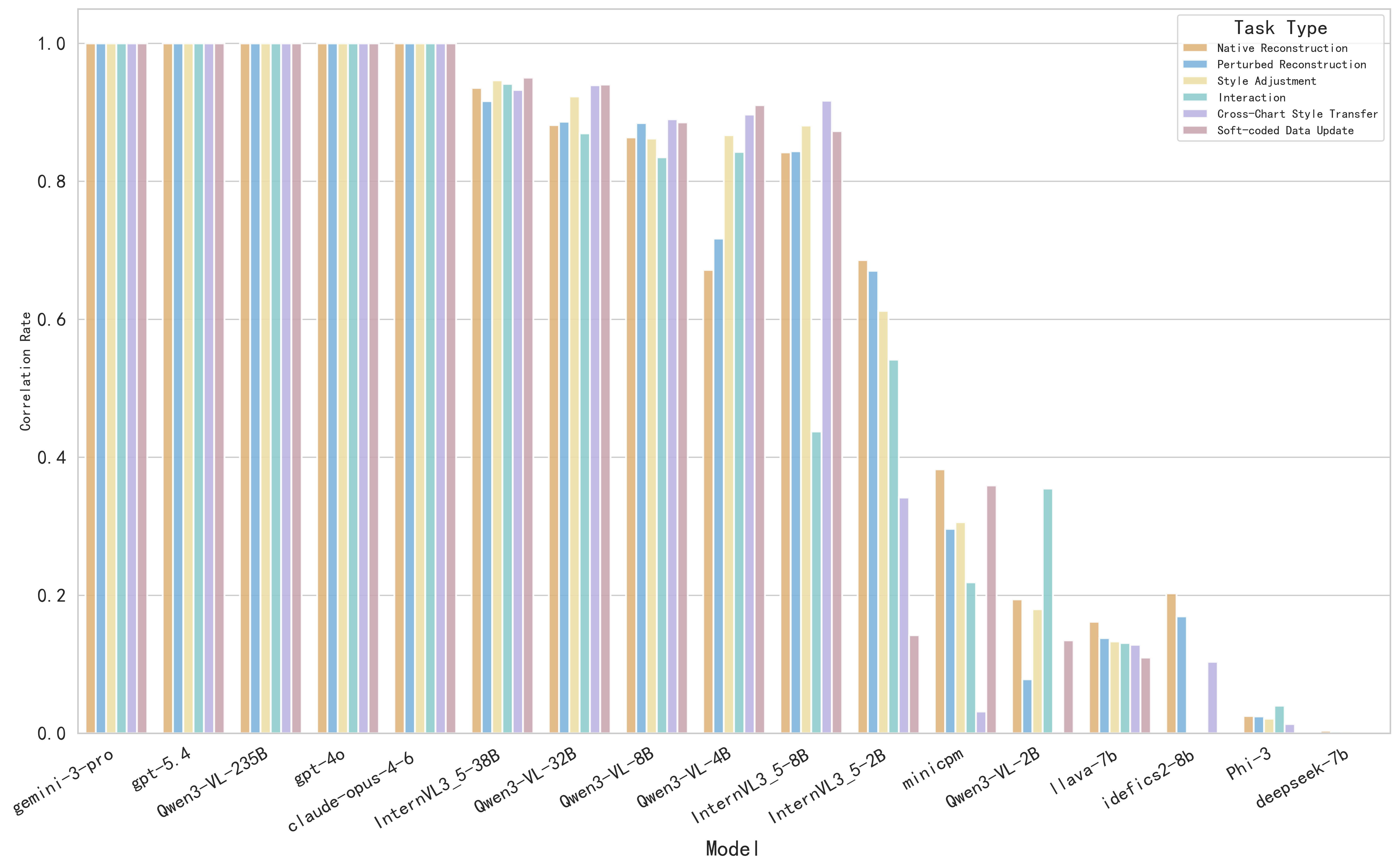}\\[-1mm]
        {\footnotesize (a) Code executability}
    \end{minipage}\hfill
    \begin{minipage}{0.49\linewidth}
        \centering
        \includegraphics[width=\linewidth, trim={20mm 20mm 15mm 0mm}, clip]{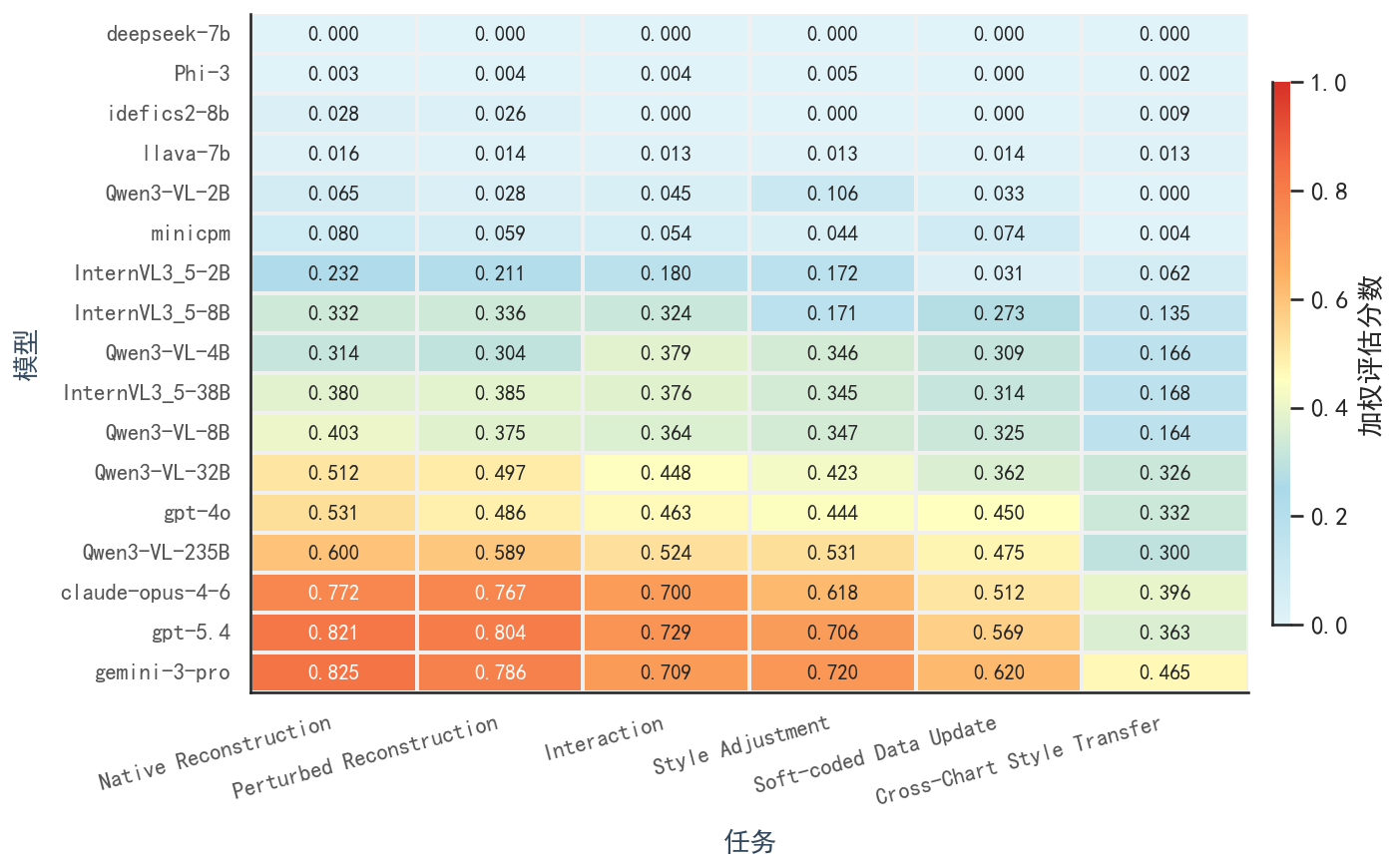}\\[-1mm]
        {\footnotesize (b) Final weighted score}
    \end{minipage}
    \caption{Functional validity and aggregate performance across ChartInt tasks.}
    \label{fig:correct_rate}
    \label{fig:hot}
    \Description{Two-panel figure comparing code executability and final weighted score across ChartInt tasks.}
\end{figure*}

We evaluate model capabilities through code executability (Figure~\ref{fig:correct_rate}), task-specific metrics (Table~\ref{tab:finegrained}), and qualitative cases (Figure~\ref{fig:model_comparison}), addressing two questions:

\begin{itemize}
    \item \textbf{RQ1: Interaction Fidelity and the Static-Dynamic Gap.}
    Can MLLMs go beyond static visual mimicry to correctly implement dynamic interaction logic? We investigate whether models can synthesize code that not only renders the initial view correctly but also responds accurately to user events (e.g., hovering, tooltips) as verified by our Preset State Snapshot Protocol. Specifically, we quantify the performance gap between static reconstruction and dynamic interaction implementation.

    \item \textbf{ RQ2: Real-World Usability and the Reproduction-Adaptability Gap.}
    Does proficiency in static reproduction translate to real-world usability? Moving beyond monotonic visual replication, we evaluate whether MLLMs can act as reliable assistants in realistic maintenance workflows. We assess model performance across Data Update, Style Adjustment, and Style Transfer to determine if they can effectively adapt existing visualizations without compromising functional integrity.
\end{itemize}

\begin{figure}[t]
    \centering
    \includegraphics[width=1\linewidth]{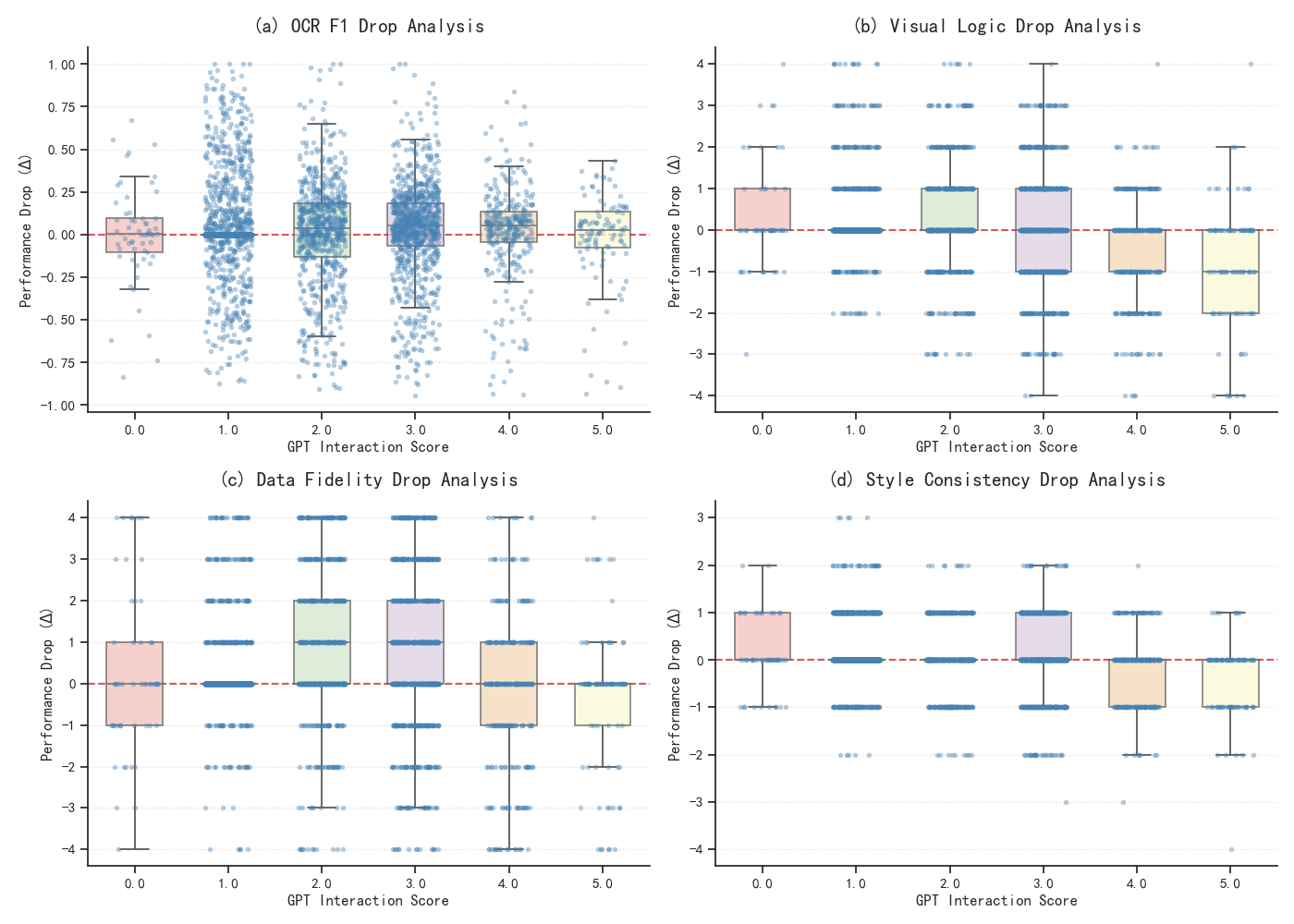}
    \caption{Correlation analysis between interaction quality and static performance degradation. The x-axis represents the interaction quality score (0–5) evaluated by GPT, while the y-axis indicates the resulting performance drop in static metrics (calculated as $S_{static} - S_{interactive}$) across (a) OCR, (b) Visual Logic, (c) Data Fidelity, and (d) Style Consistency.}
    \label{fig:polarization_analysis}
    \Description{Scatter plots showing the relationship between interaction quality scores and degradation in static metrics.}
\end{figure}

\subsection{Experimental Setup}
\label{sec:exp_setup}

We evaluate open-source and proprietary MLLMs spanning lightweight to large-scale architectures (Table~\ref{tab:finegrained}). Unless otherwise specified, all models use a unified prompting and execution protocol~\cite{wei2022chain}. This comparison assesses chart reproduction, interaction, and editing across model capacities.

\subsection{RQ1: Interaction Fidelity and the Static-Dynamic Gap}
To answer RQ1, we trace model capabilities across three increasing levels of complexity: Native, Perturbed, and Interaction. By contrasting these tasks, we isolate whether performance drops stem from visual noise or fundamental deficits in dynamic logic.

\subsubsection{Static Robustness vs. Dynamic Fragility.}
Leading models (e.g., Gemini-3-Pro) demonstrate remarkable resilience to static visual complexity. As shown in Table \ref{tab:finegrained}, although the Perturbed Reconstruction task introduces complex stylistic variations that make visual reproduction more challenging, models suffer only negligible performance degradation compared to the native baseline. In stark contrast, the Interaction task—which operates on the visually simpler native charts but demands dynamic logic—triggers a significantly steeper performance decline.
This control experiment isolates the root cause of failure: the bottleneck is not visual recognition difficulty, but the model's fundamental inability to handle dynamic state transitions. This finding highlights that interaction capabilities are a critical, yet currently overlooked, dimension of chart understanding that cannot be measured by static visual complexity alone.

\begin{figure*}[t]
    \centering
    \setlength{\tabcolsep}{2pt}
    \renewcommand{\arraystretch}{1.5}

    \begin{adjustbox}{max width=\linewidth,max totalheight=0.80\textheight}
\begin{tabular}{r cccccc}
        & \textbf{Native} & \textbf{Perturbed} & \textbf{Interaction} & \textbf{Style Adjustment} & \textbf{Data Update} & \textbf{Style Transfer} \\
        
        \rotatebox{90}{\textbf{GT}} & \includegraphics[width=0.145\textwidth]{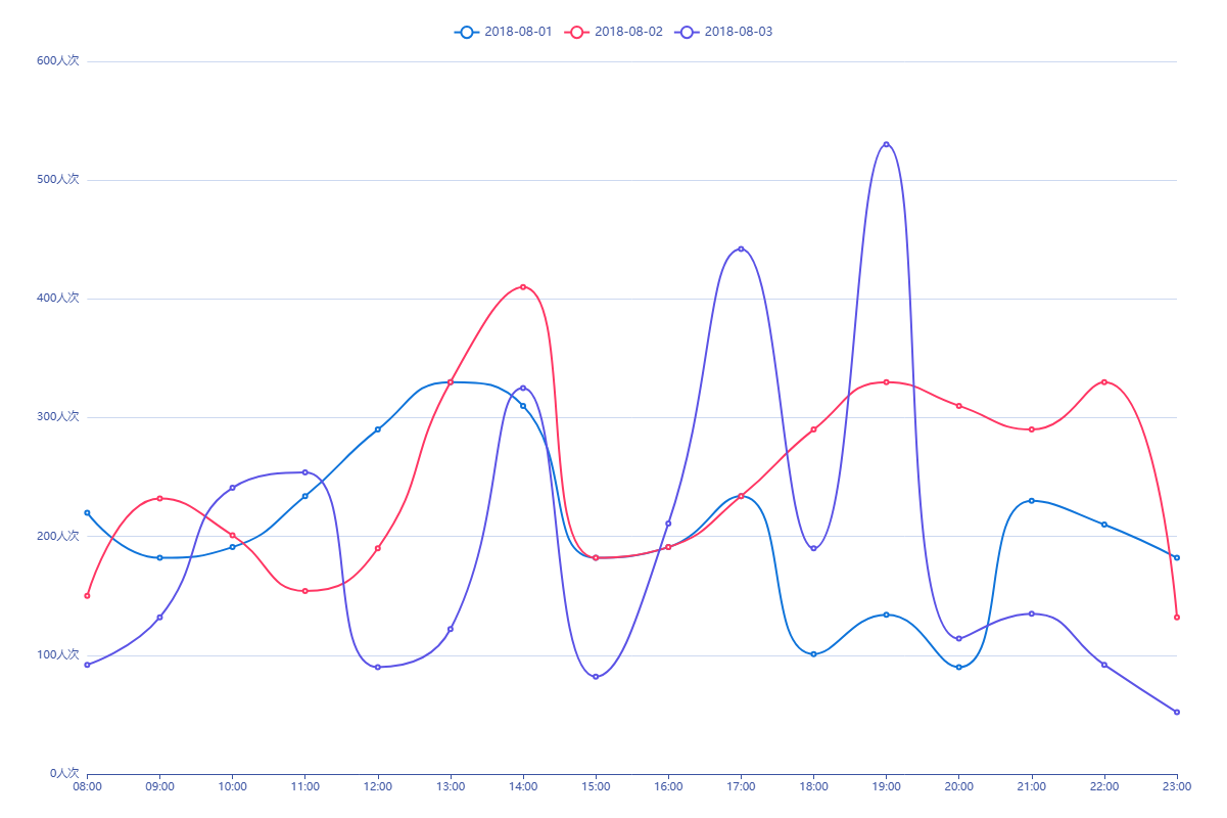} & \includegraphics[width=0.145\textwidth]{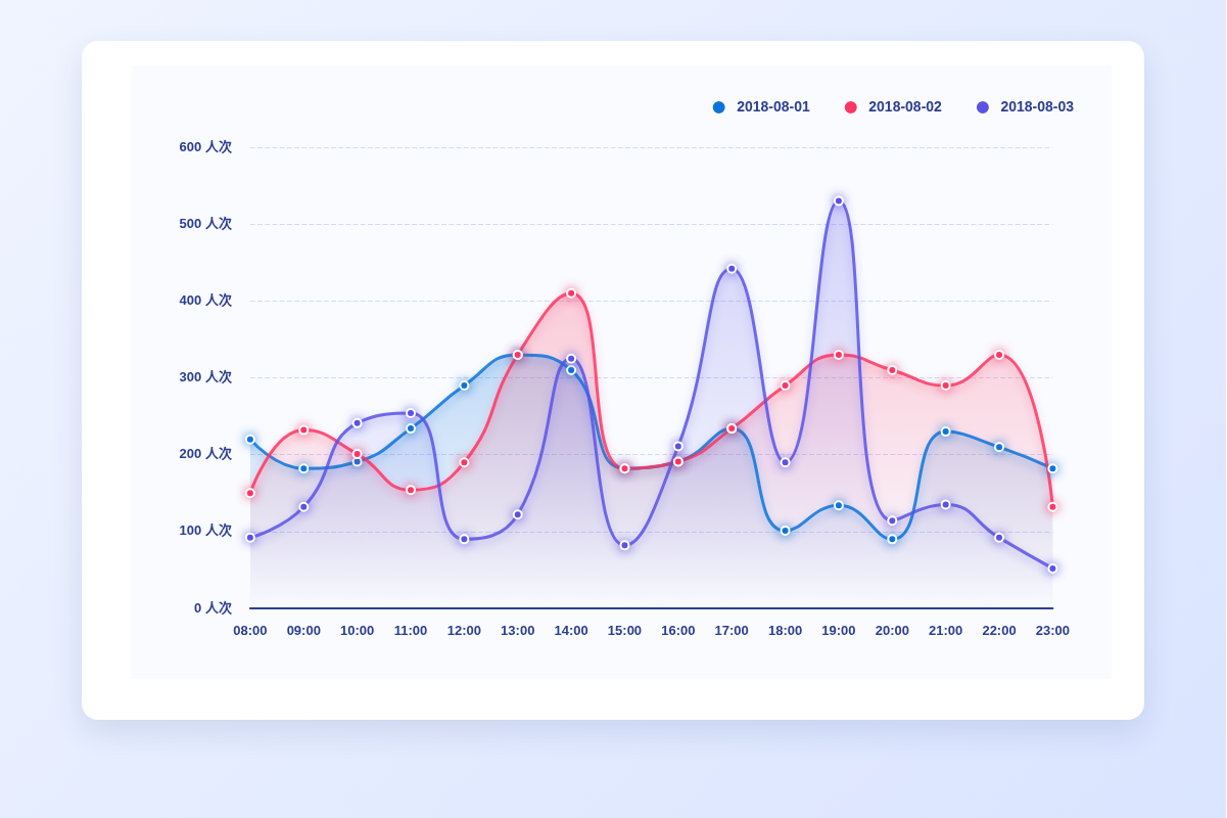} & \includegraphics[width=0.145\textwidth]{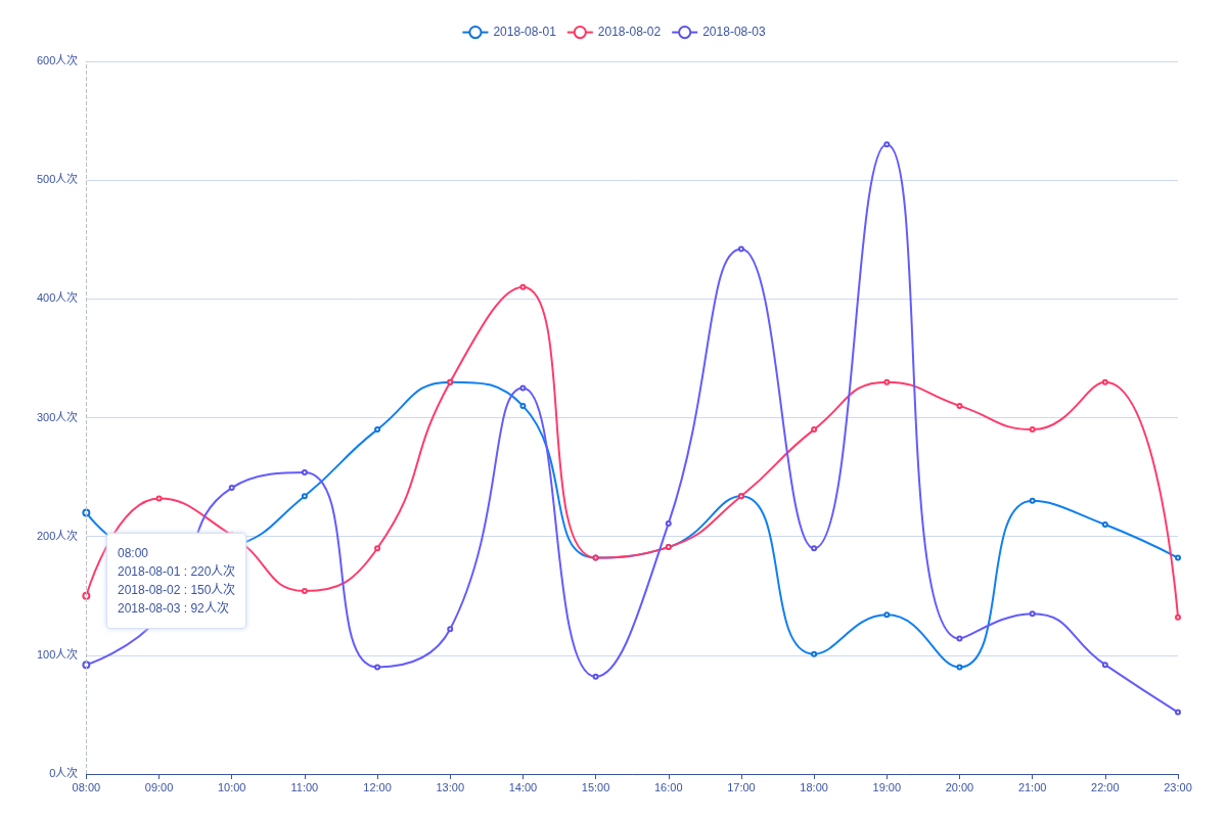} & \includegraphics[width=0.145\textwidth]{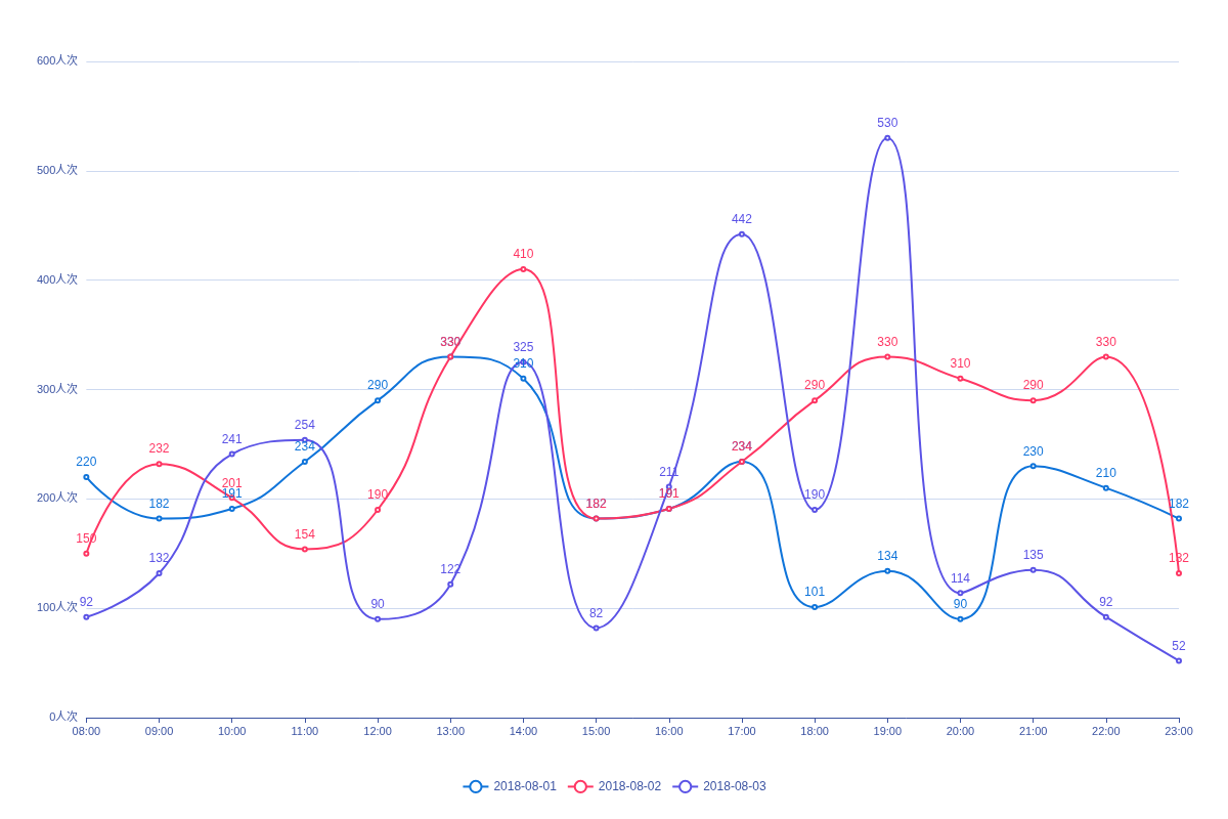} & \includegraphics[width=0.145\textwidth]{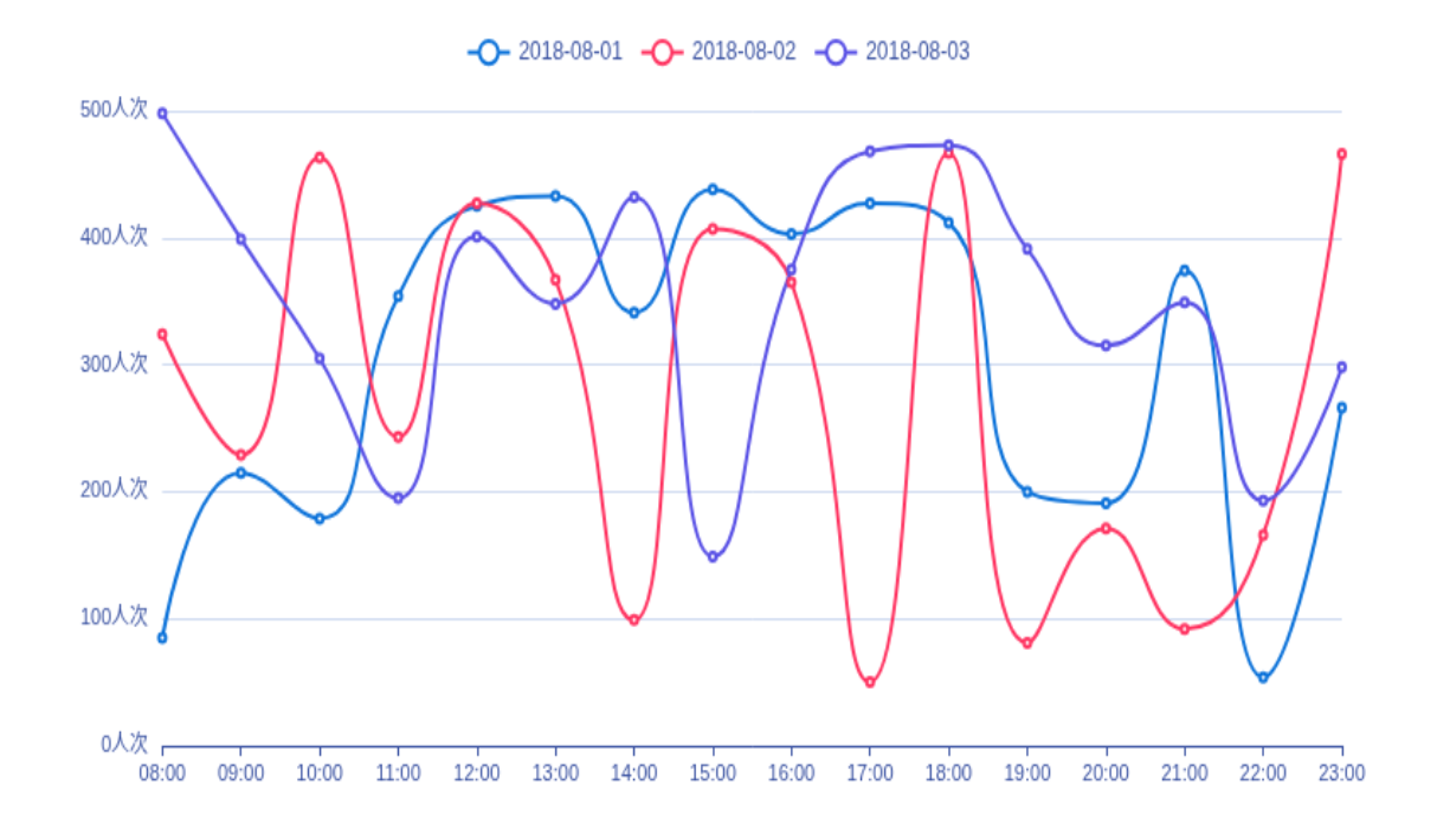} & \includegraphics[width=0.145\textwidth]{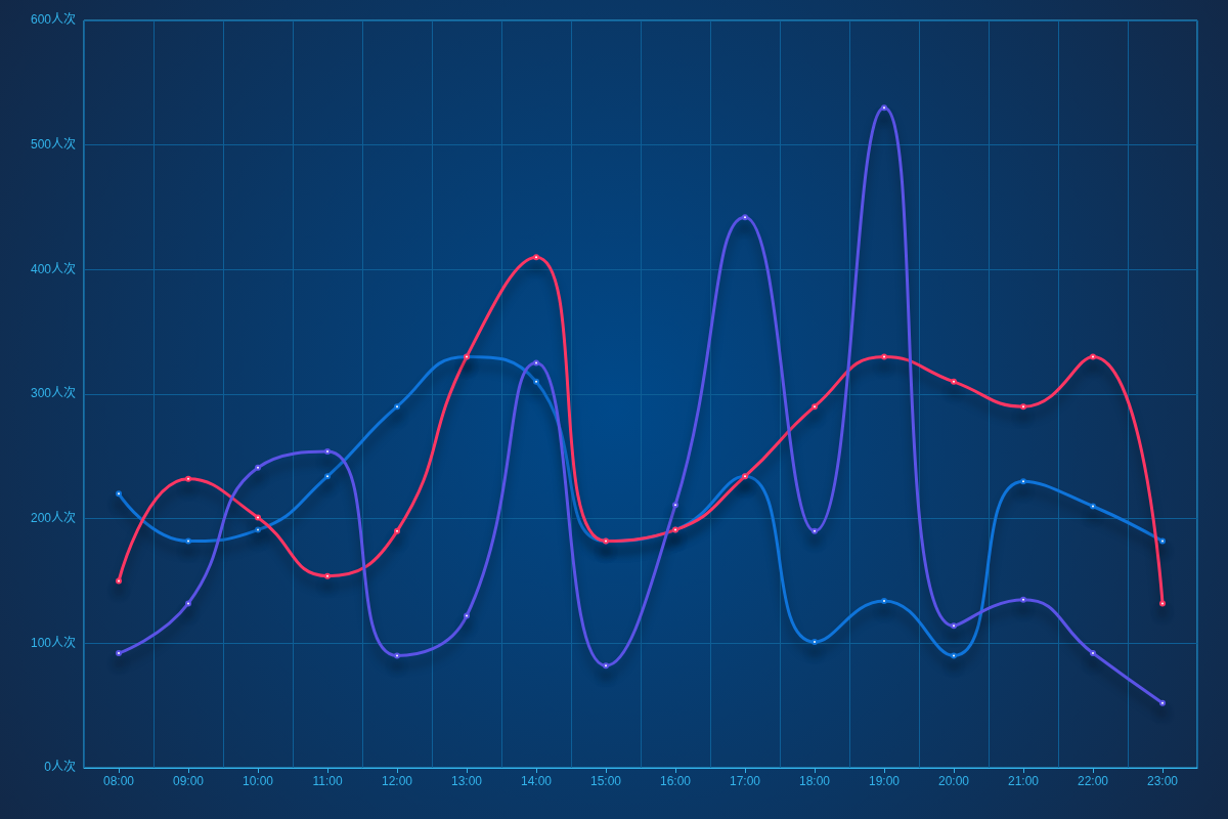} \\
        
        \rotatebox{90}{\textbf{Gemini}} & \includegraphics[width=0.145\textwidth]{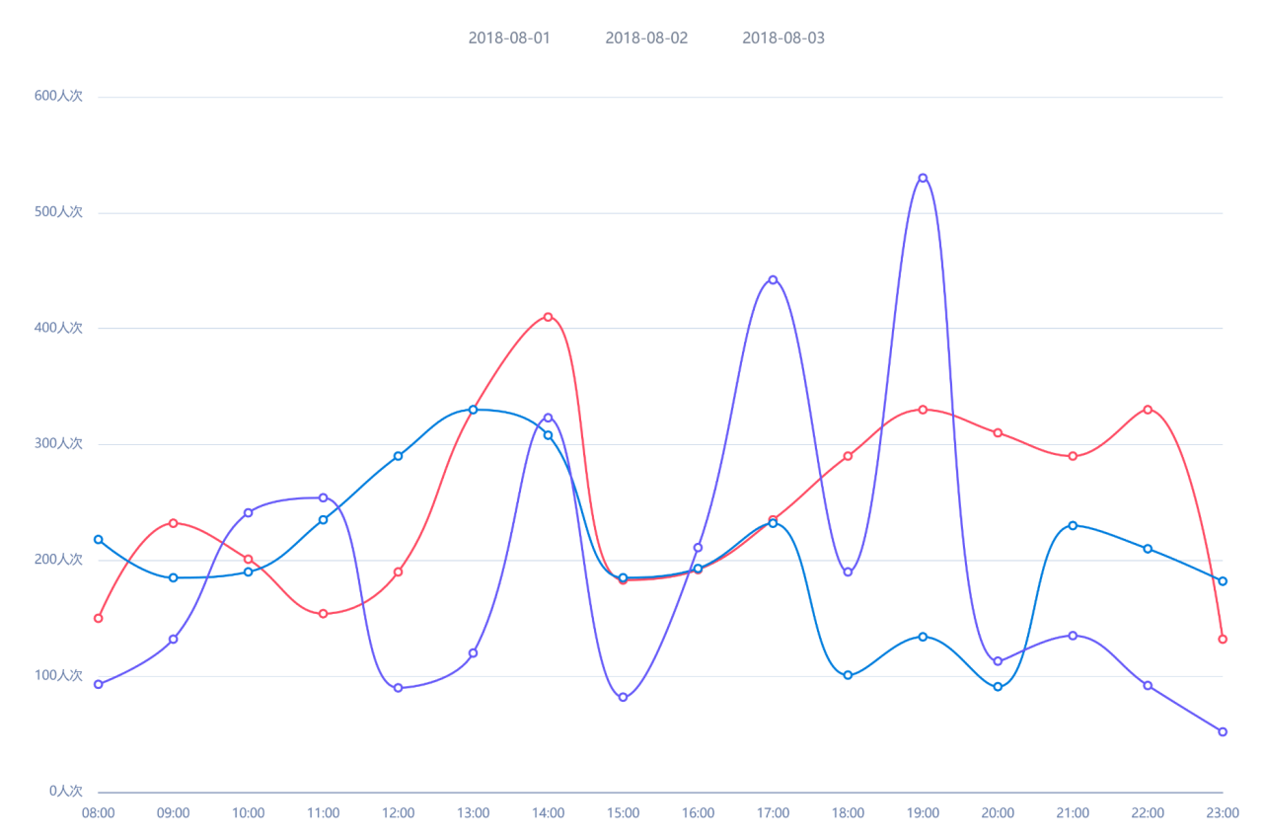} & \includegraphics[width=0.145\textwidth]{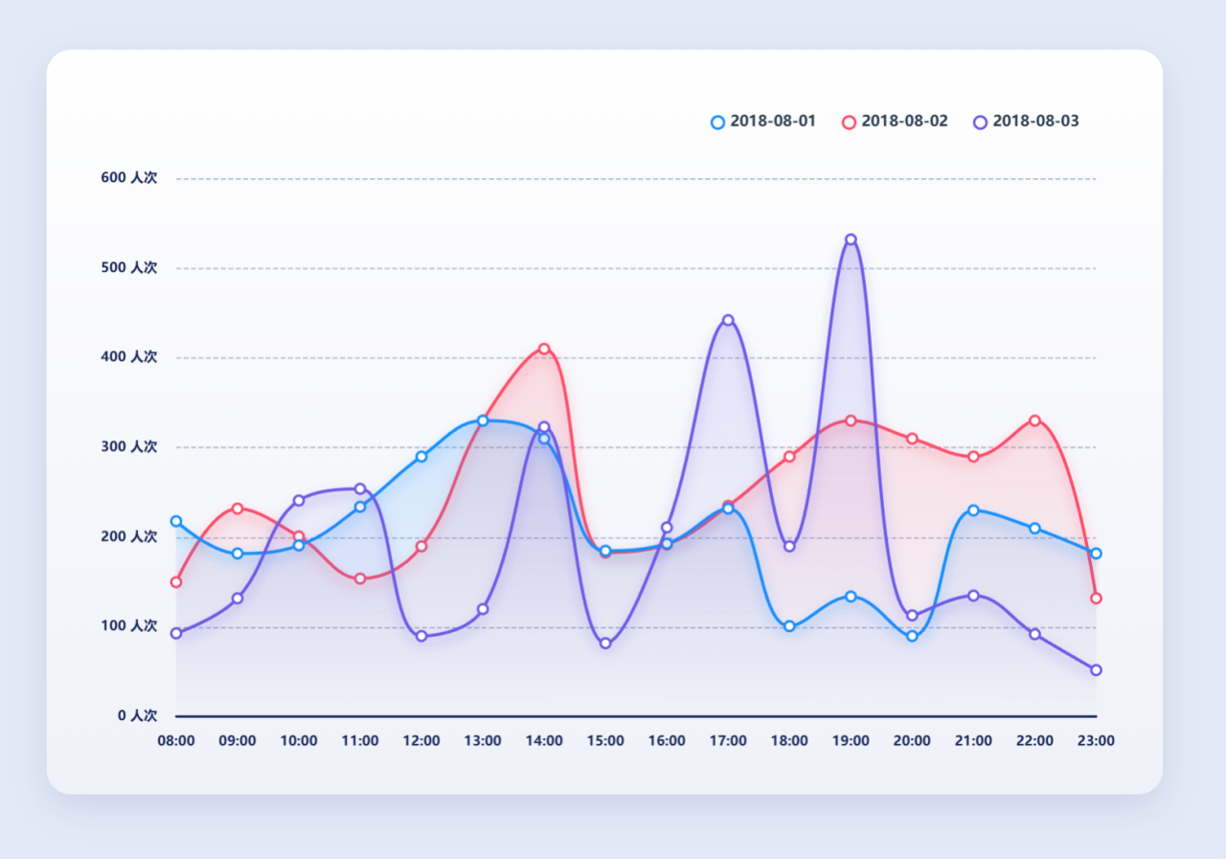} & \includegraphics[width=0.145\textwidth]{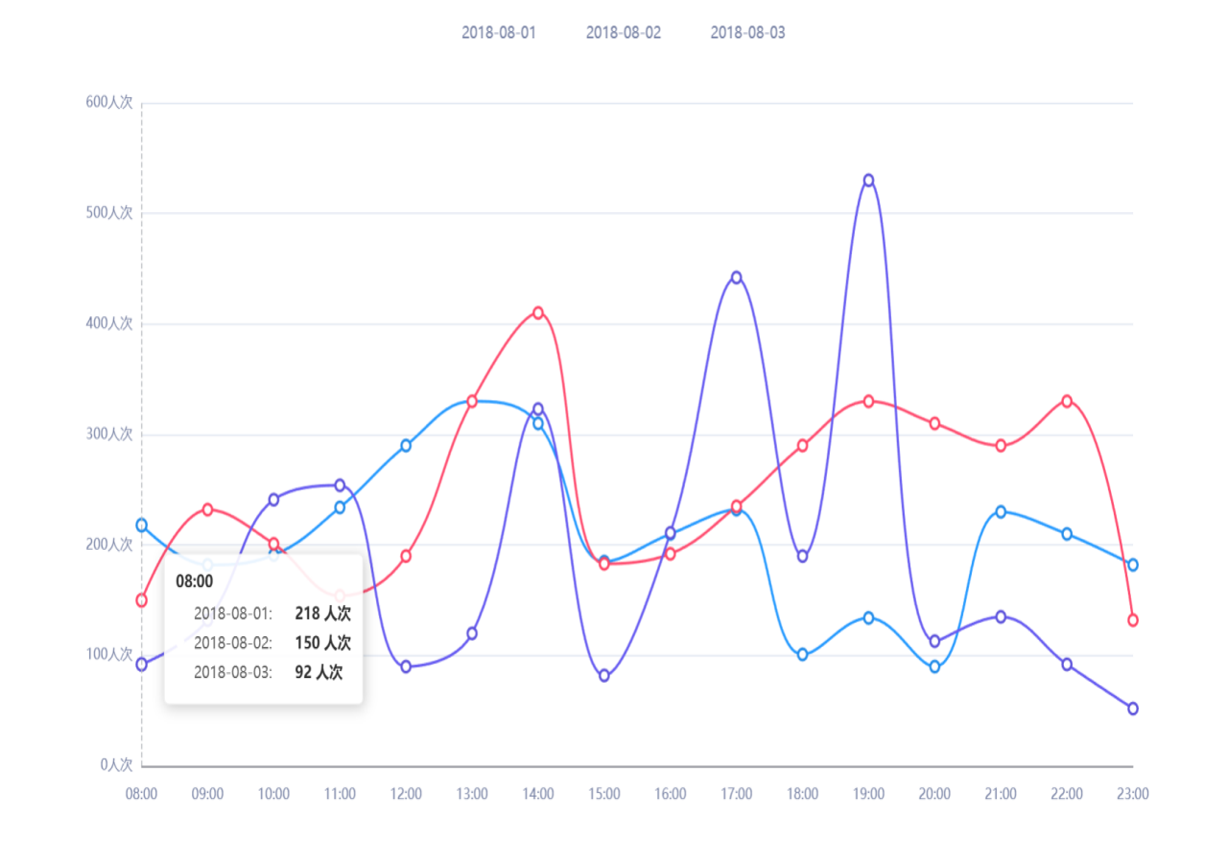} & \includegraphics[width=0.145\textwidth]{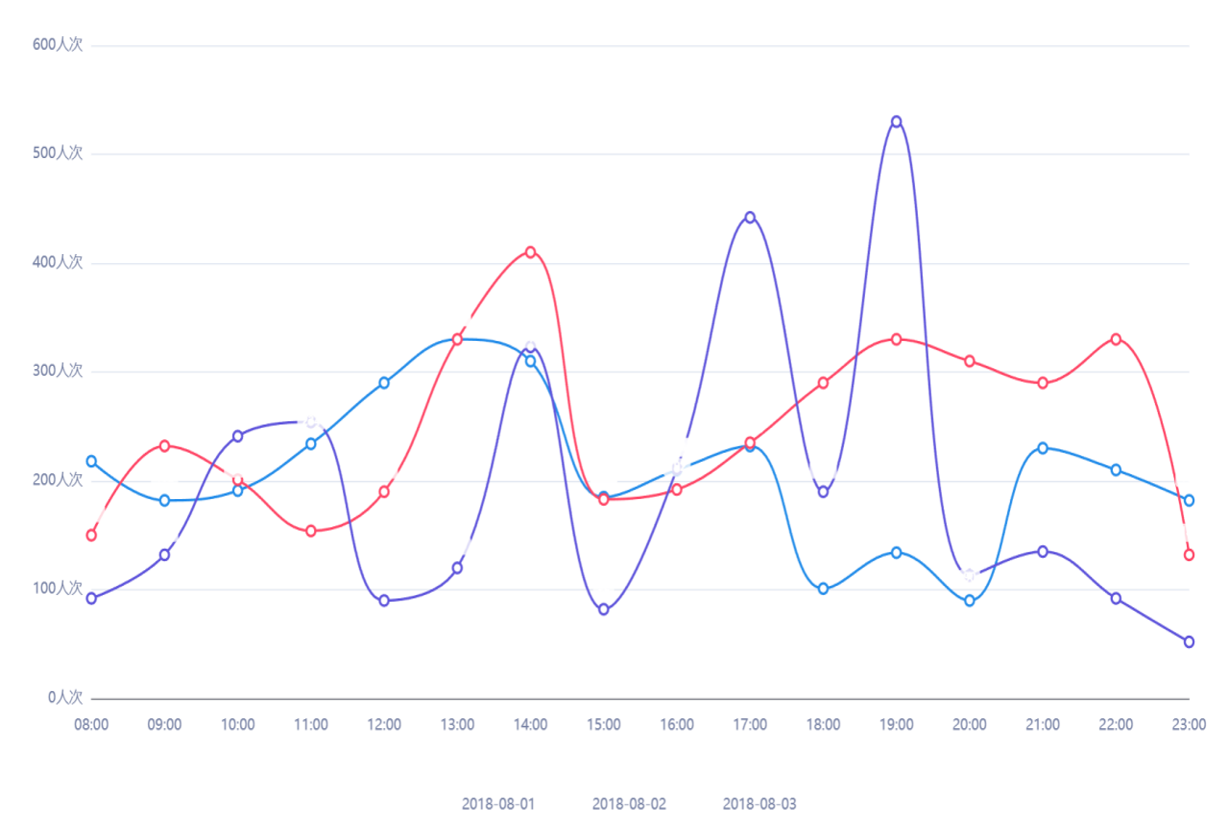} & \includegraphics[width=0.145\textwidth]{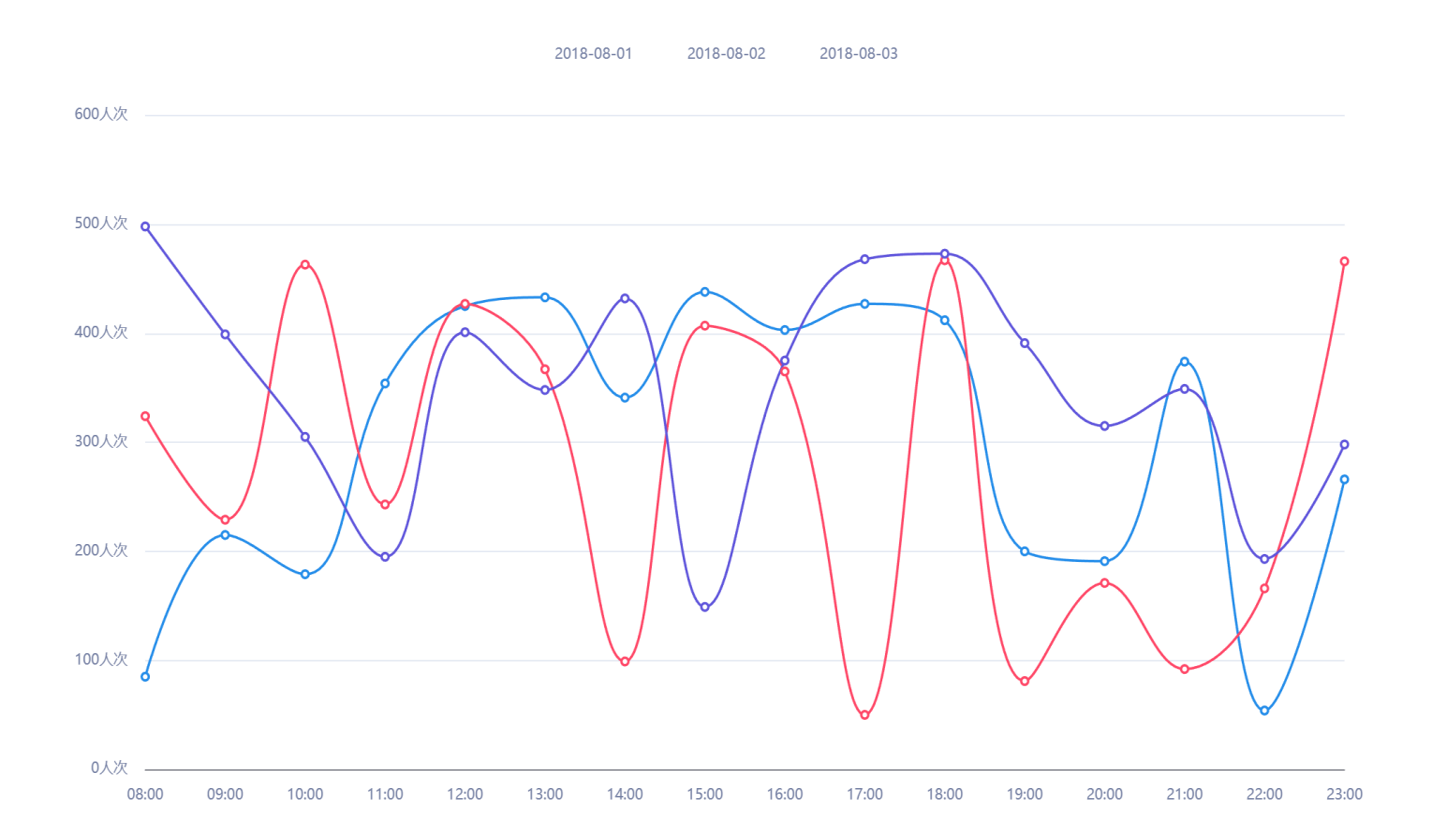} & \includegraphics[width=0.145\textwidth]{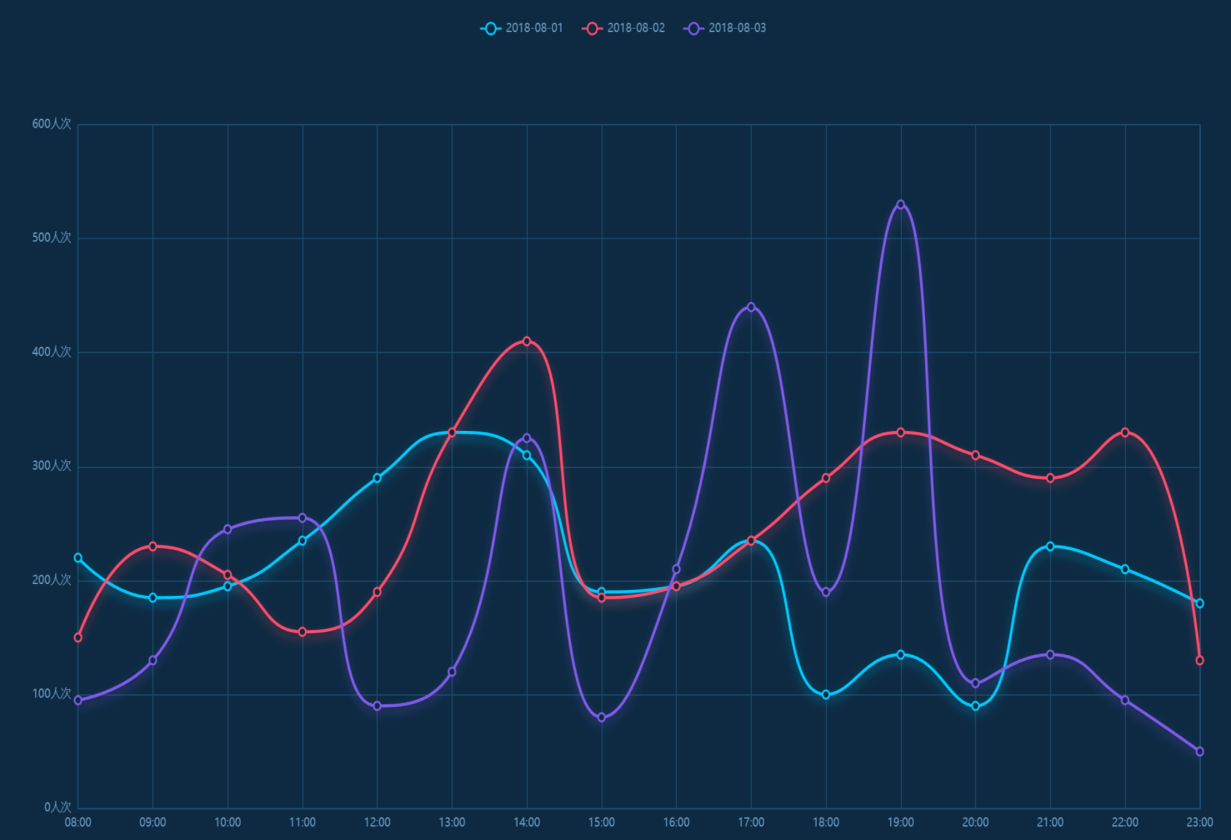} \\
        
        \rotatebox{90}{\textbf{GPT-4o}} & \includegraphics[width=0.145\textwidth]{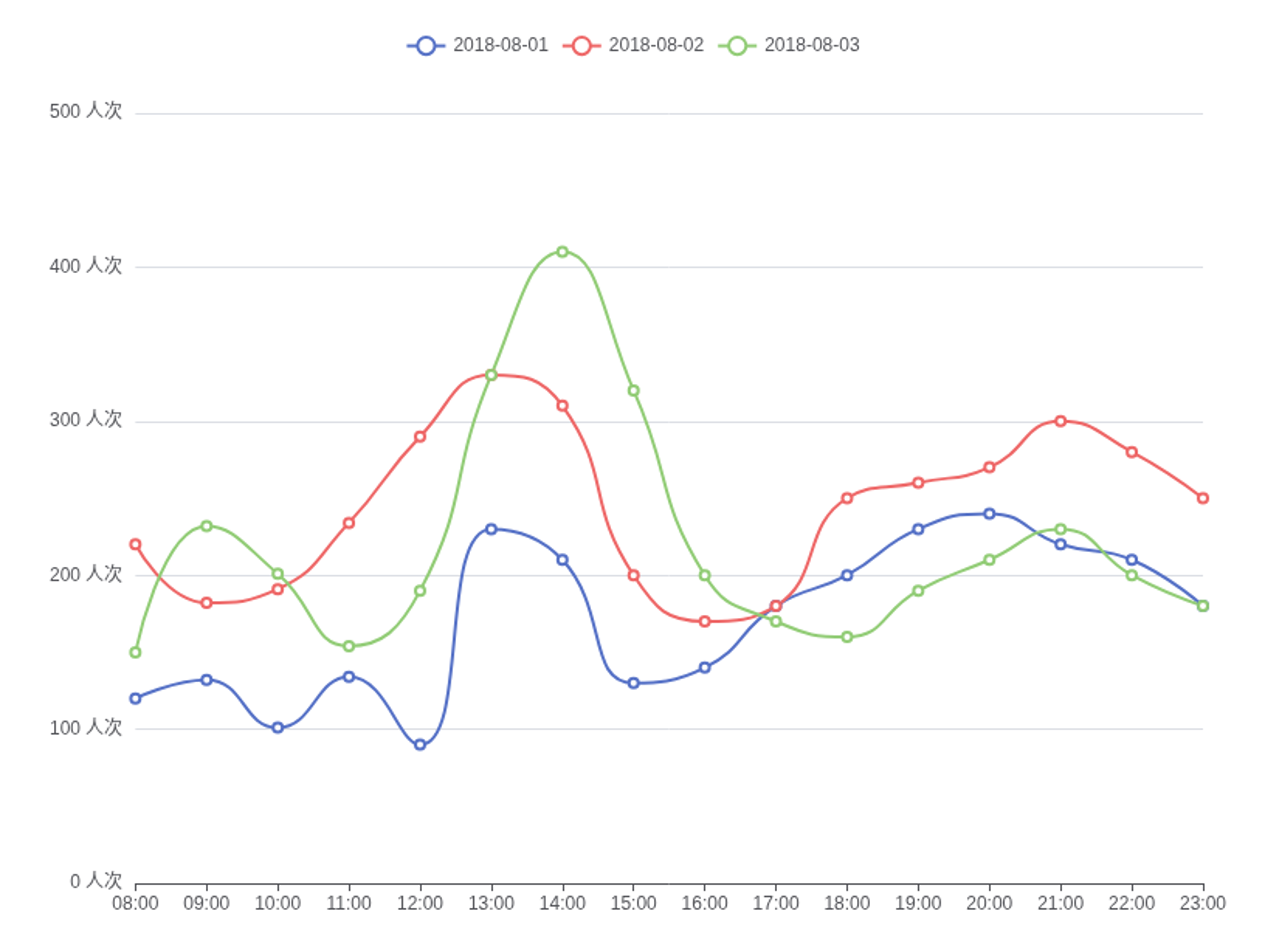} & \includegraphics[width=0.145\textwidth]{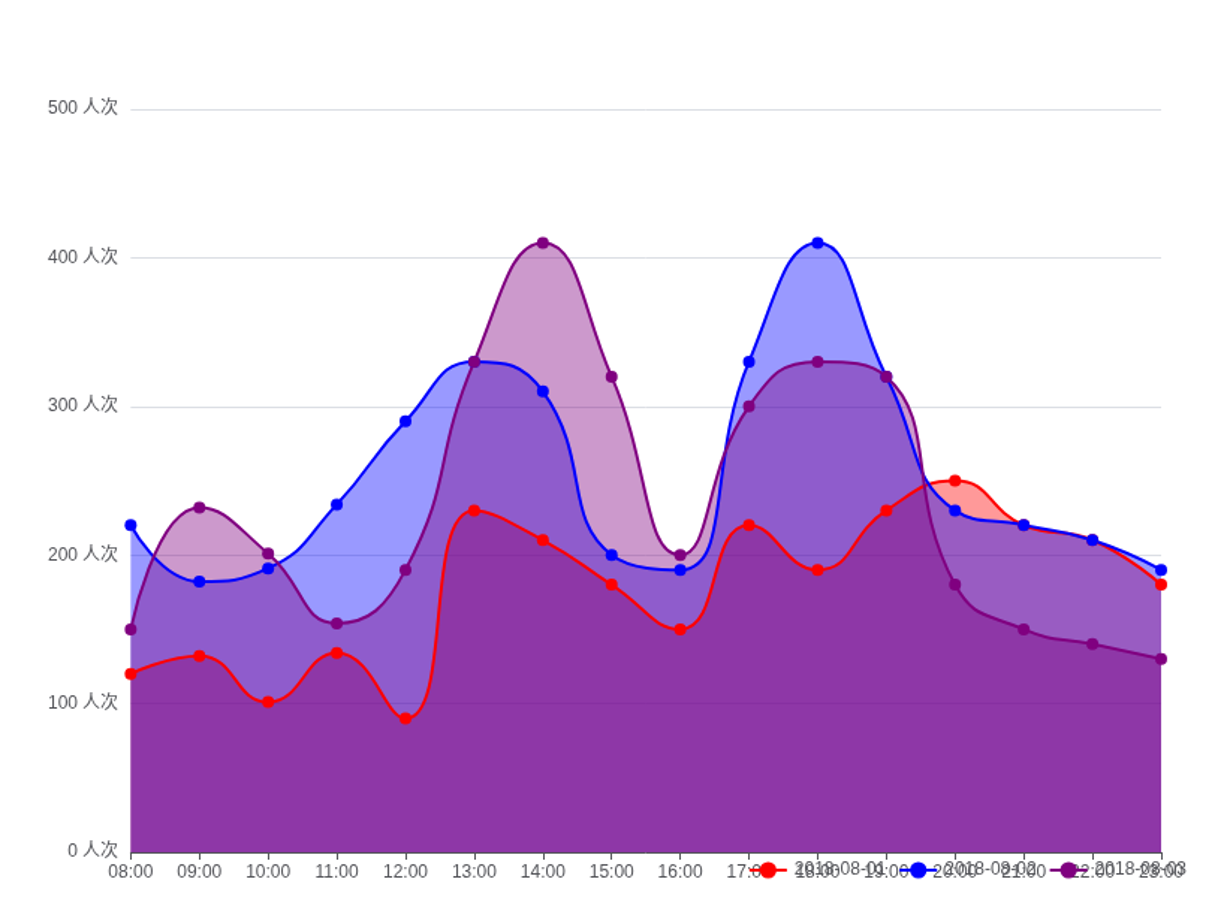} & \includegraphics[width=0.145\textwidth]{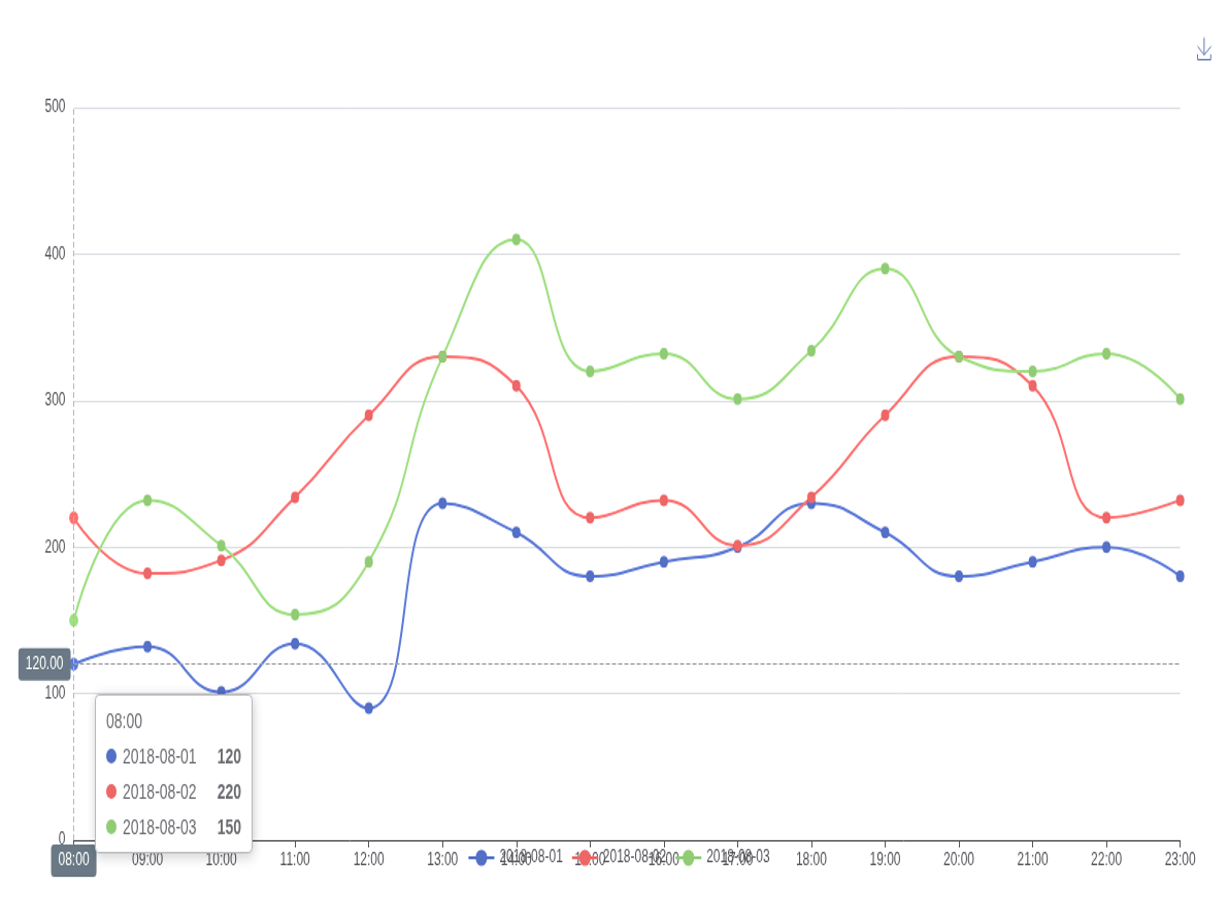} & \includegraphics[width=0.145\textwidth]{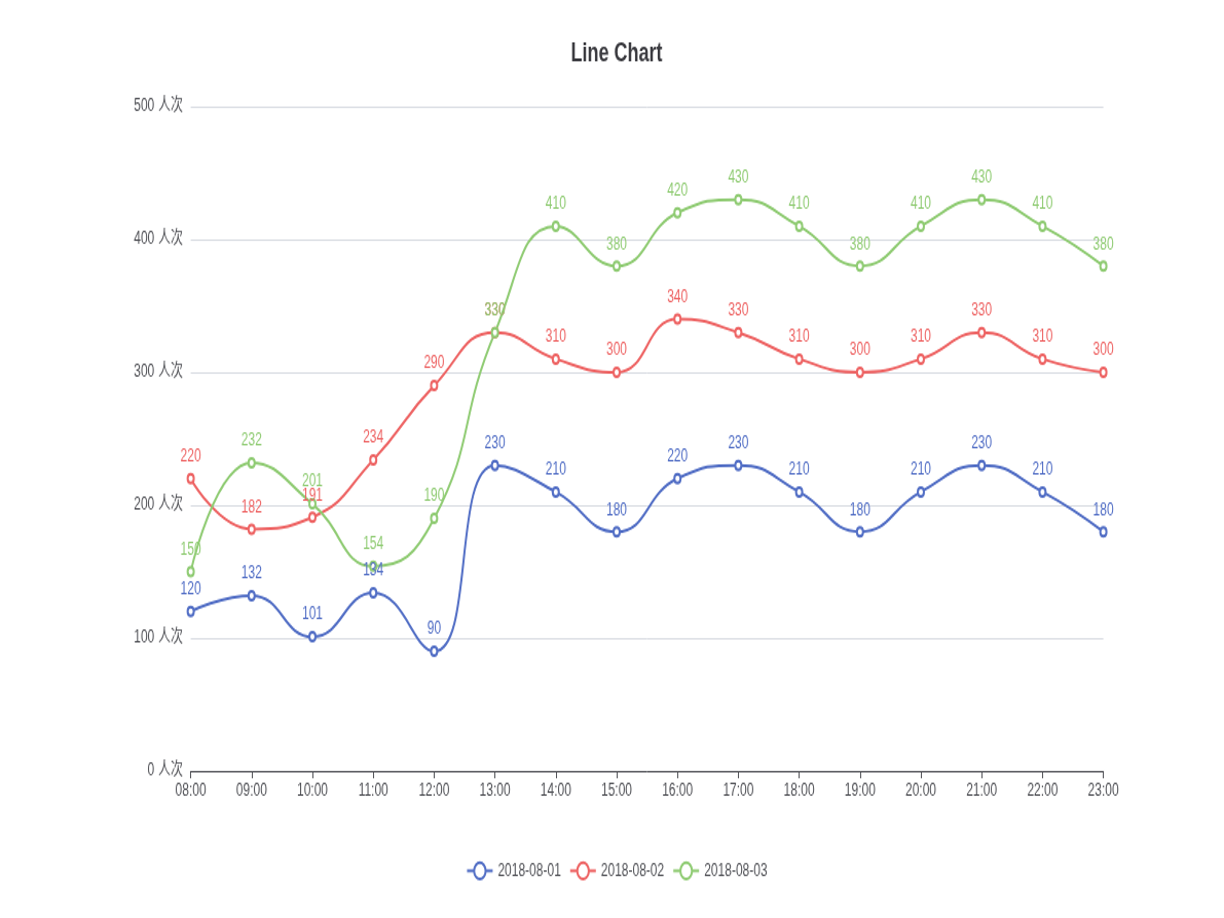} & \includegraphics[width=0.145\textwidth]{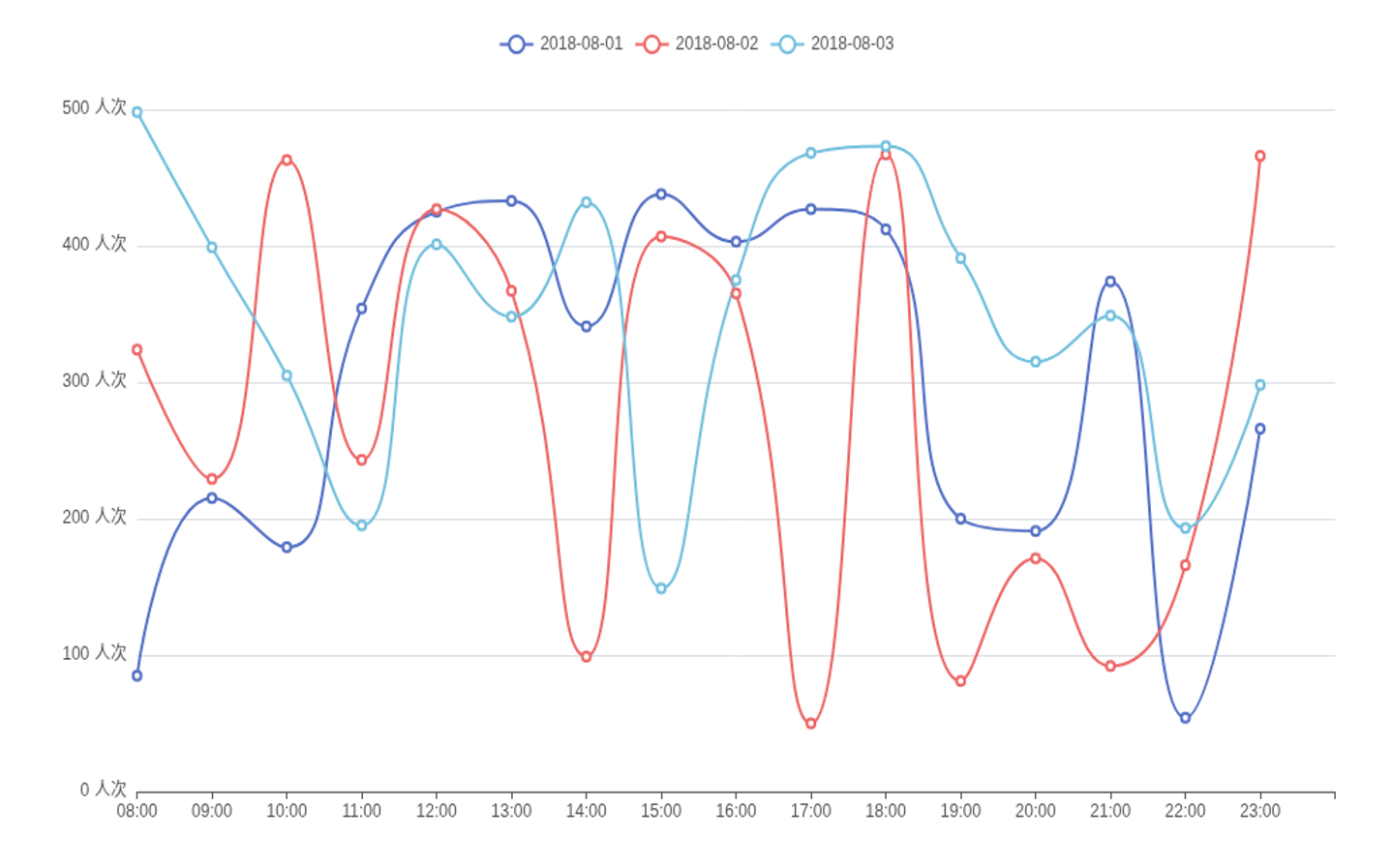} & \includegraphics[width=0.145\textwidth]{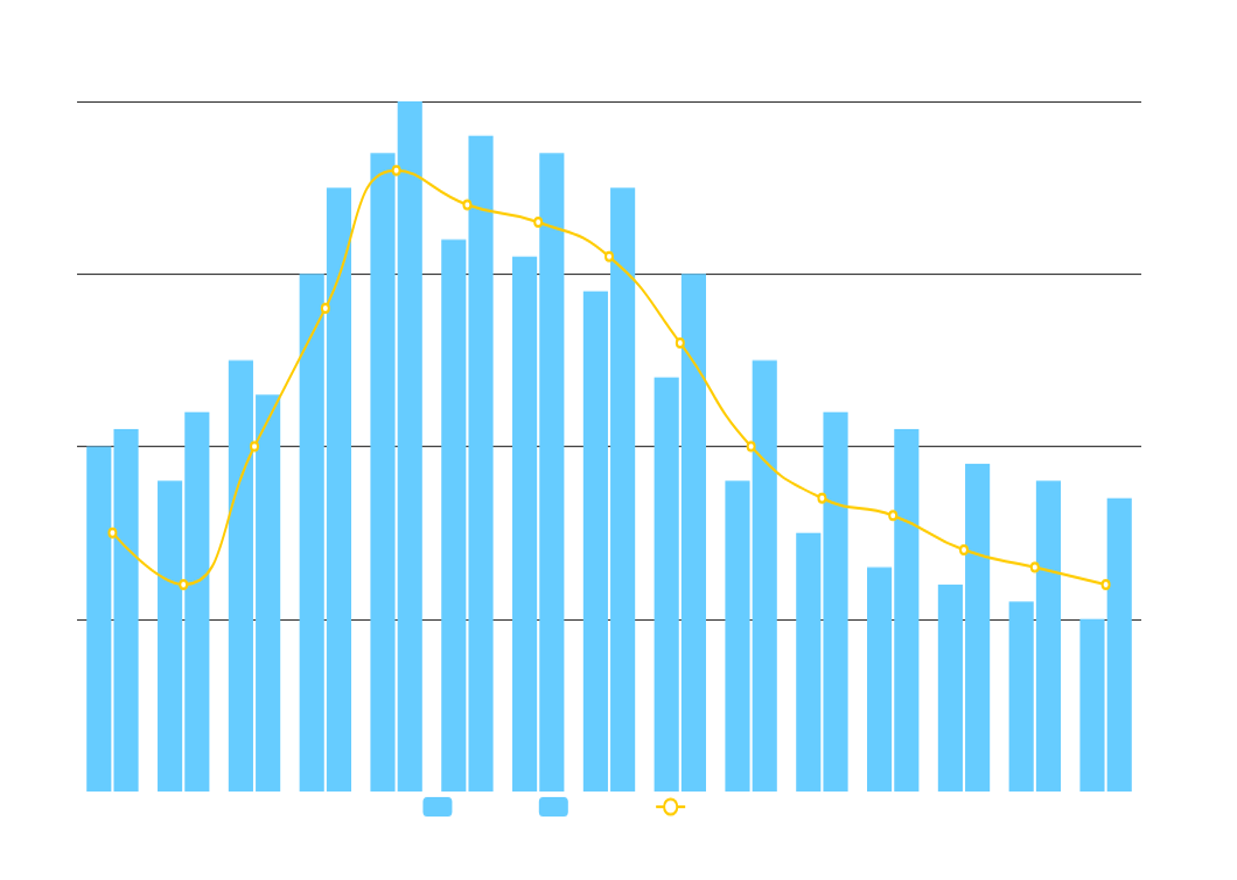} \\
        
        \rotatebox{90}{\textbf{Qwen3-32B}} & \includegraphics[width=0.145\textwidth]{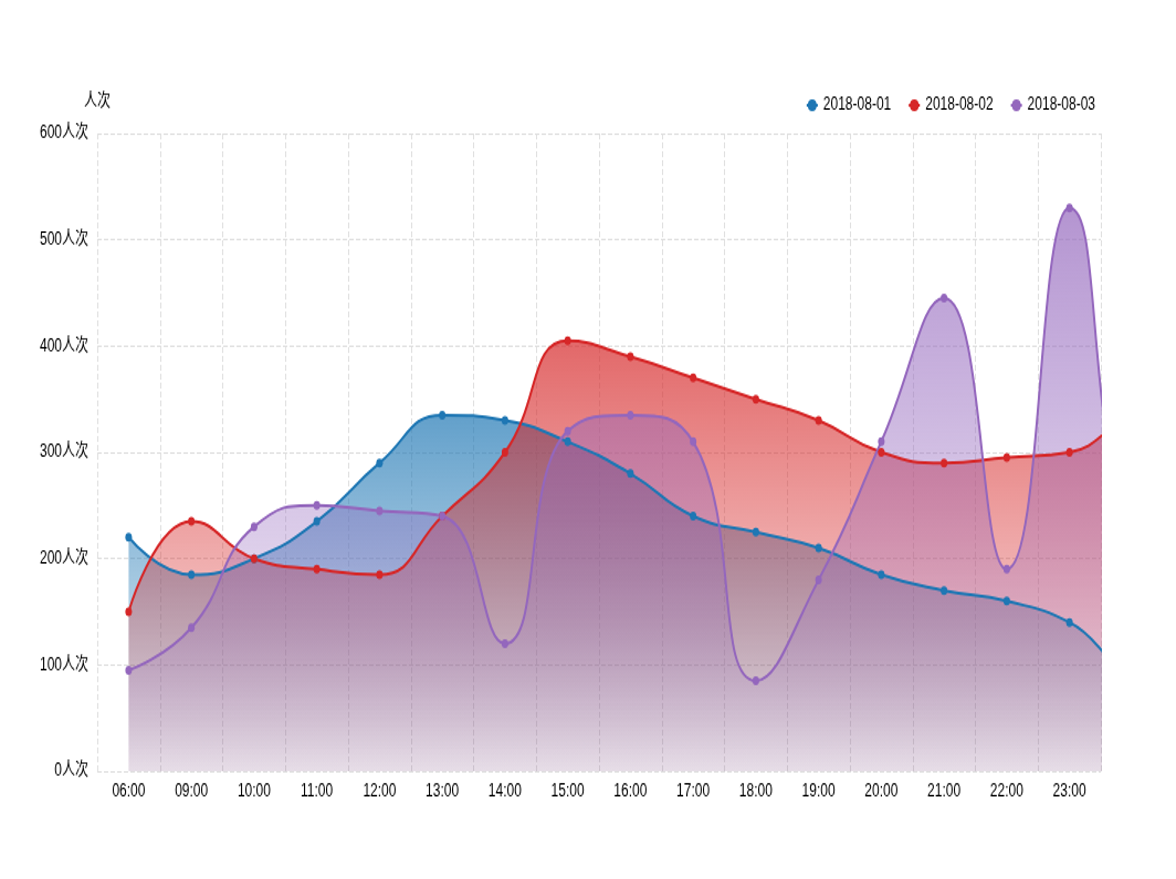} & \includegraphics[width=0.145\textwidth]{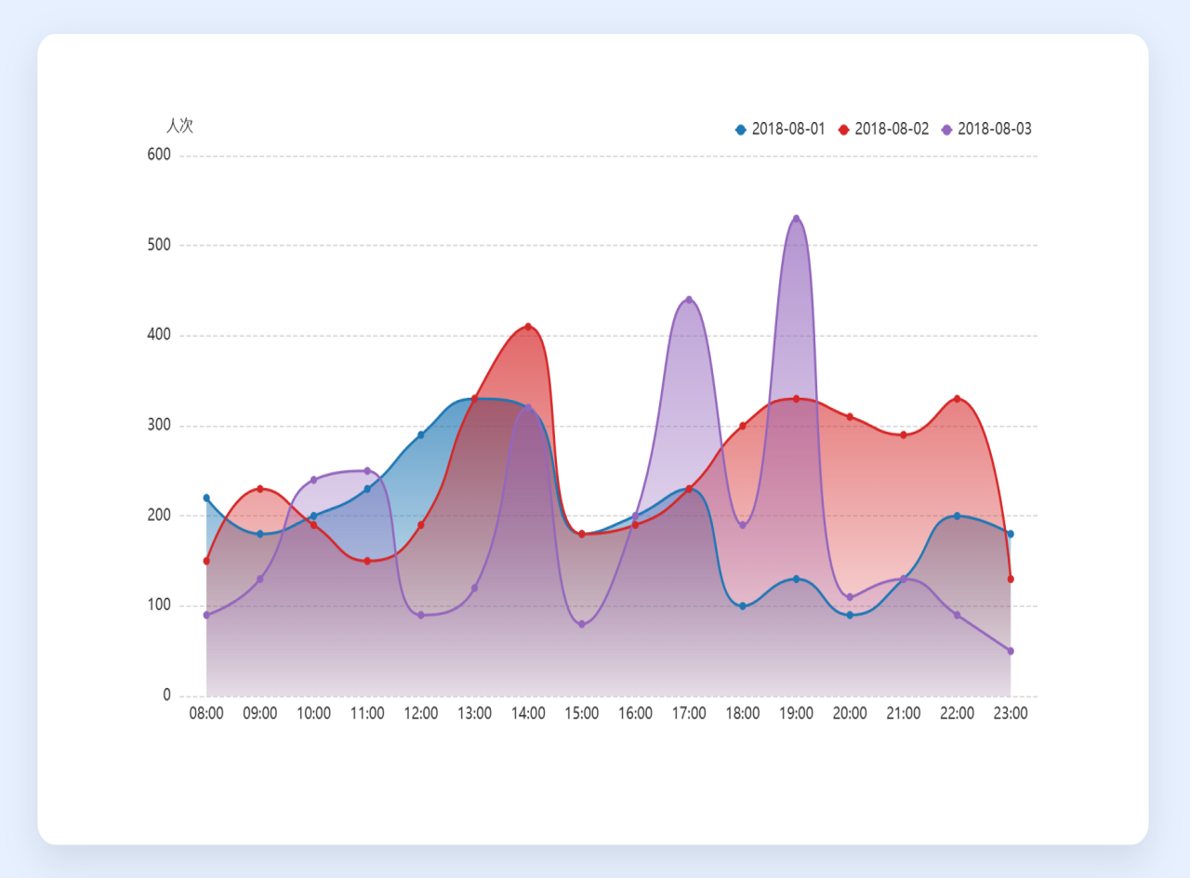} & \includegraphics[width=0.145\textwidth]{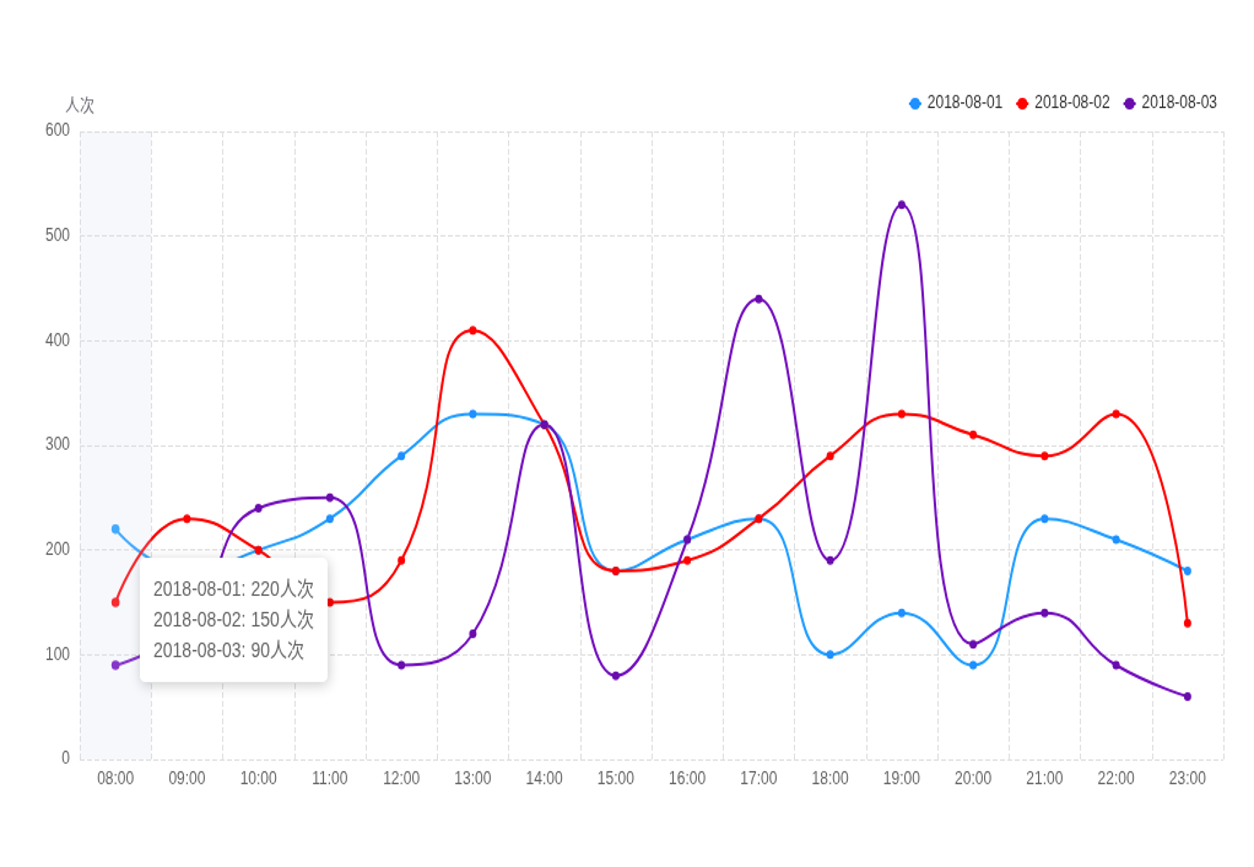} & \includegraphics[width=0.145\textwidth]{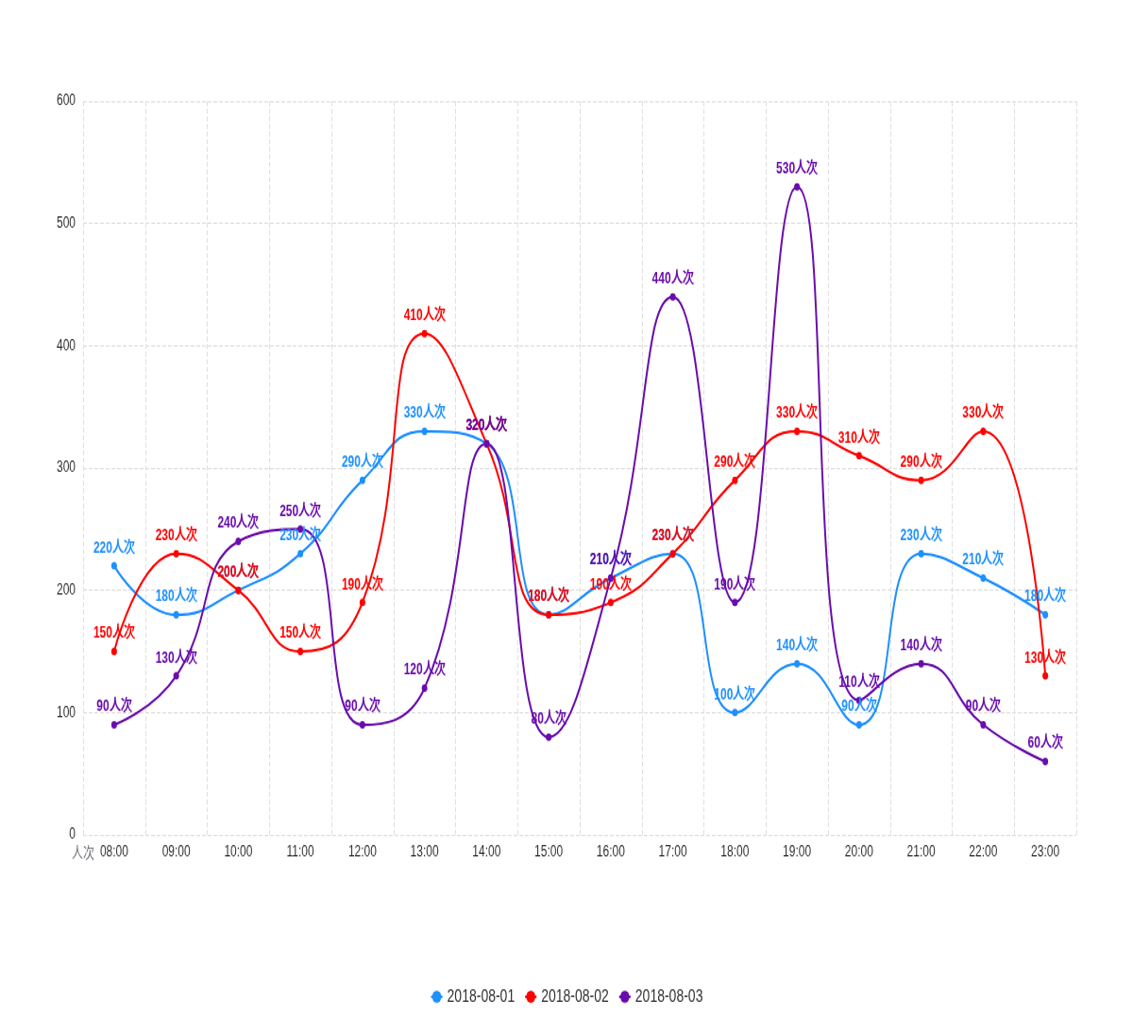} & \includegraphics[width=0.145\textwidth]{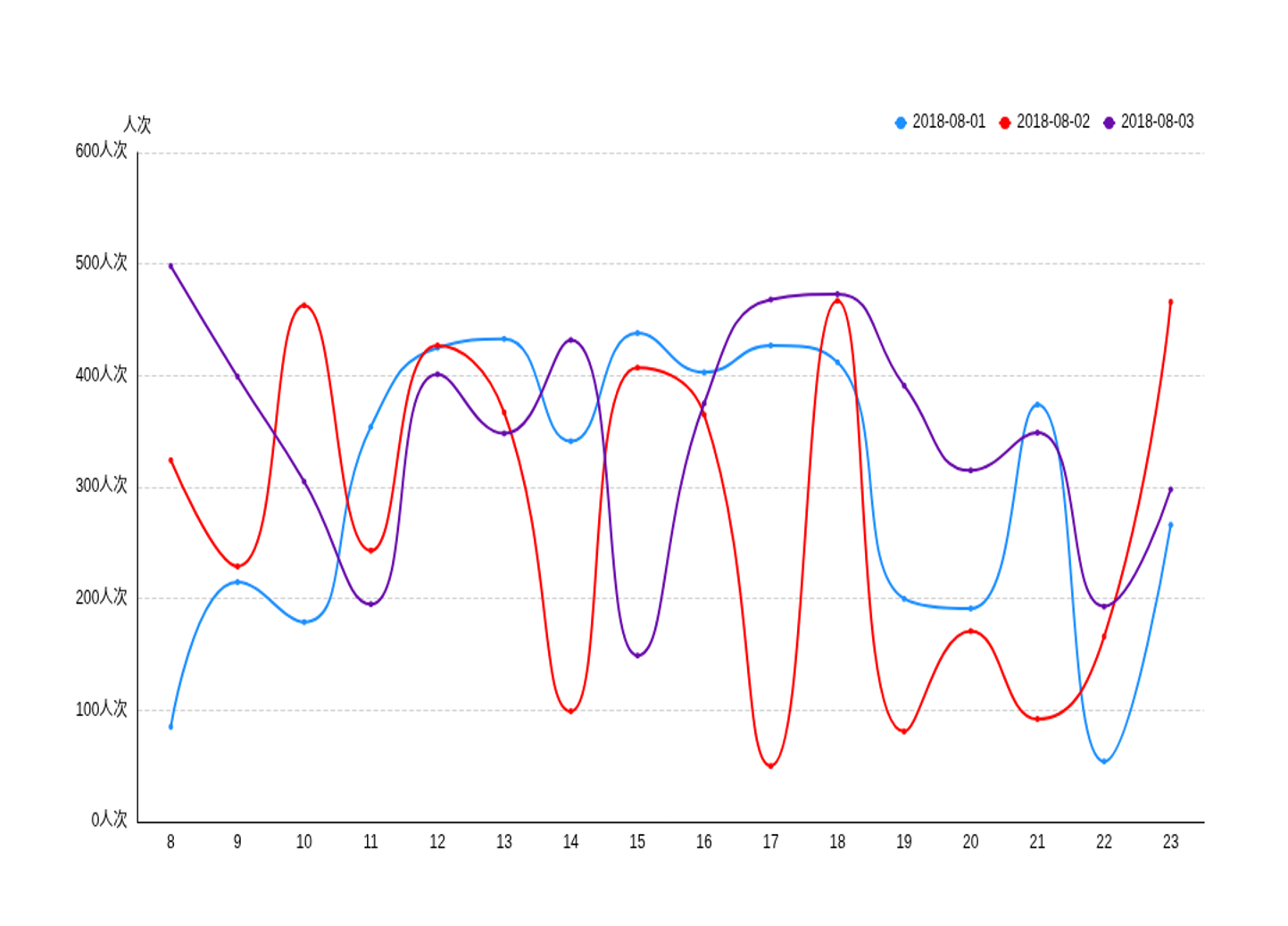} & \includegraphics[width=0.145\textwidth]{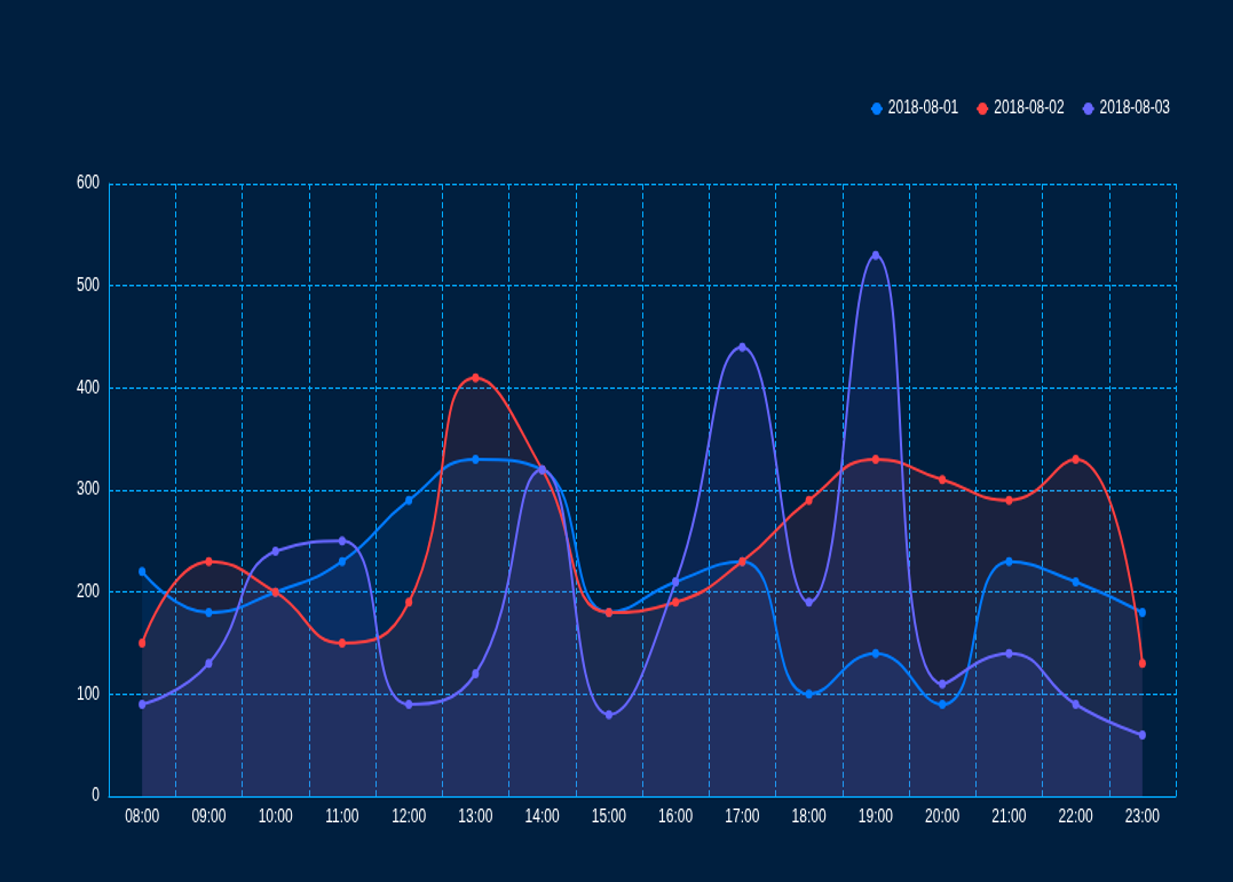} \\
        
        \rotatebox{90}{\textbf{In.3\_5-2B}} & \includegraphics[width=0.145\textwidth]{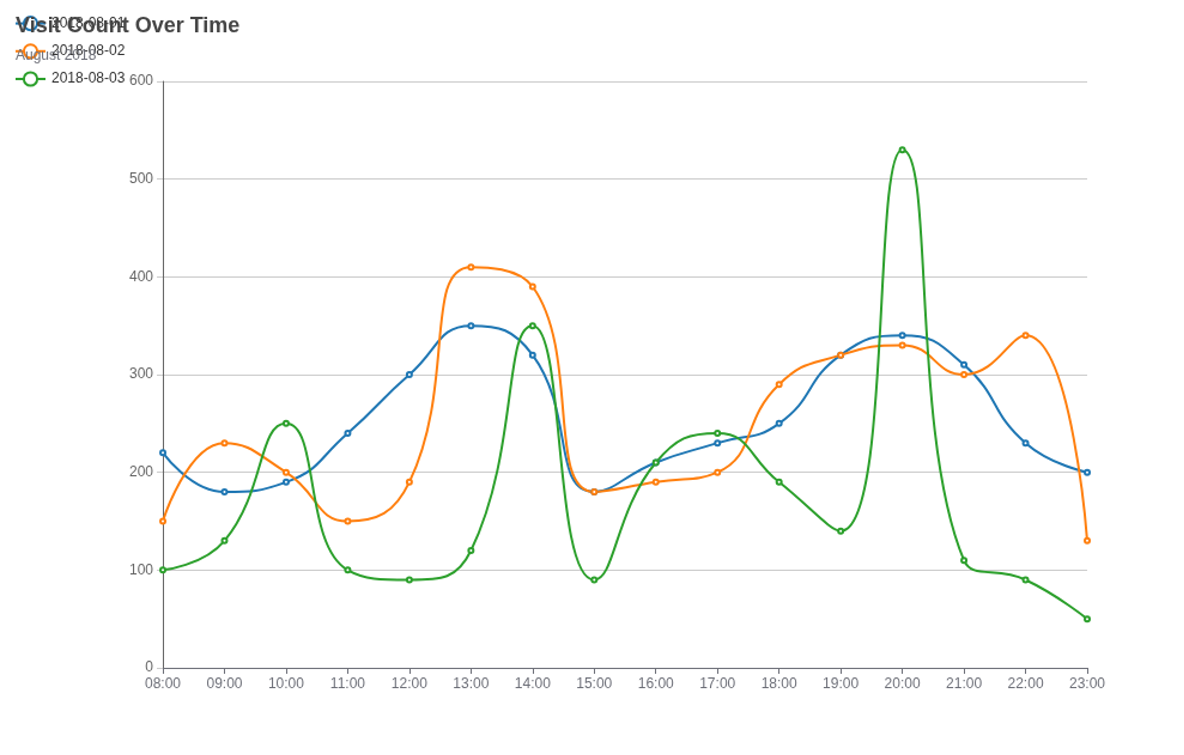} & \includegraphics[width=0.145\textwidth]{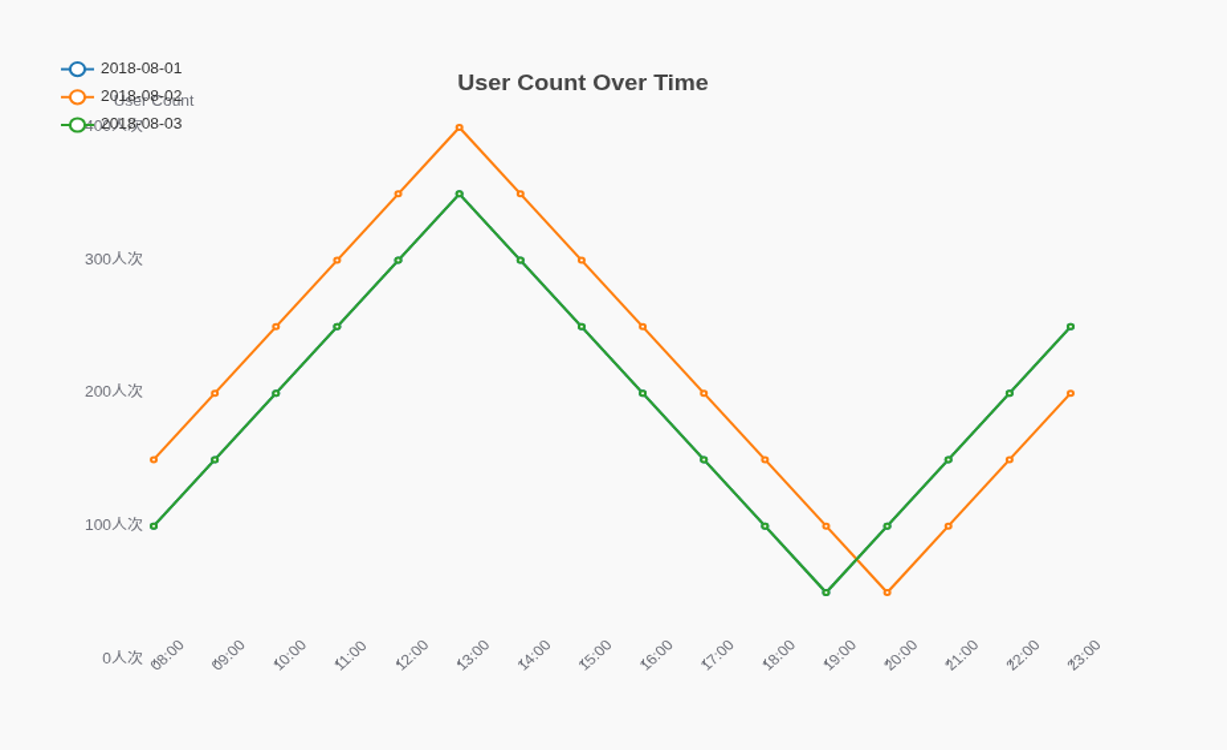} & \includegraphics[width=0.145\textwidth]{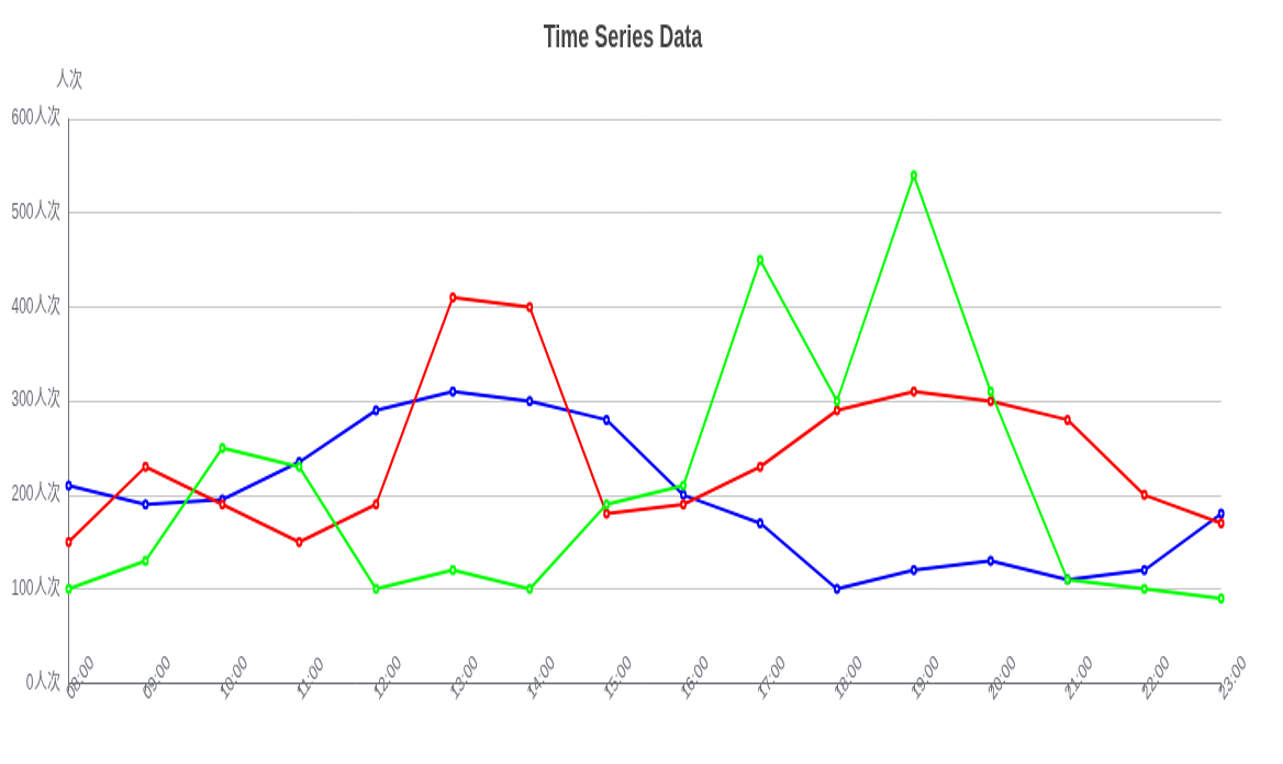} & \includegraphics[width=0.145\textwidth]{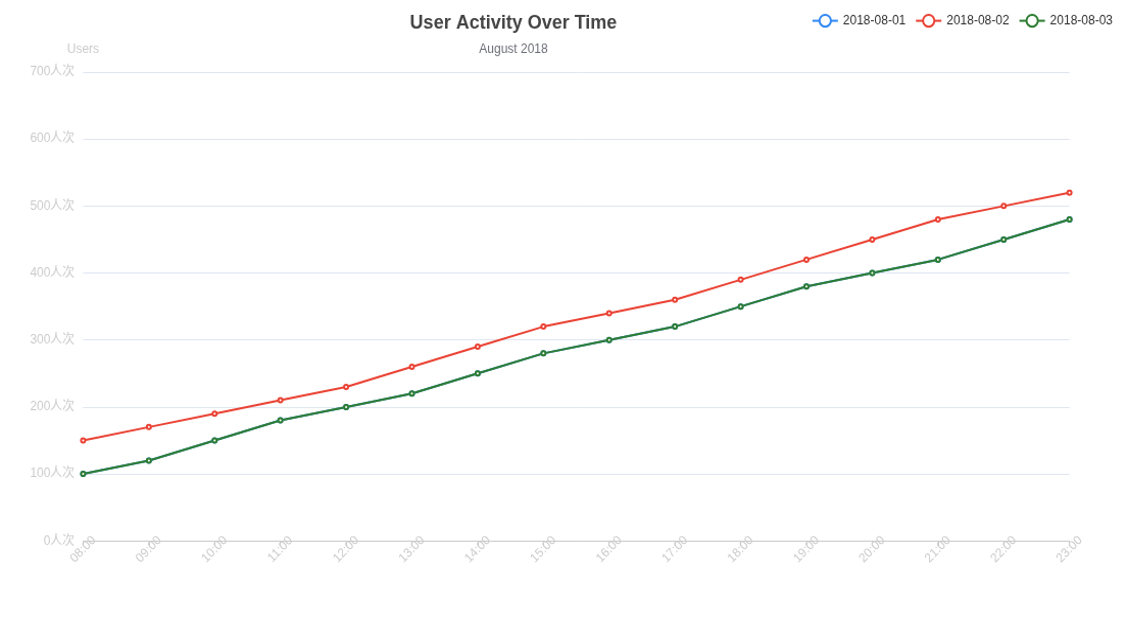} & \includegraphics[width=0.145\textwidth]{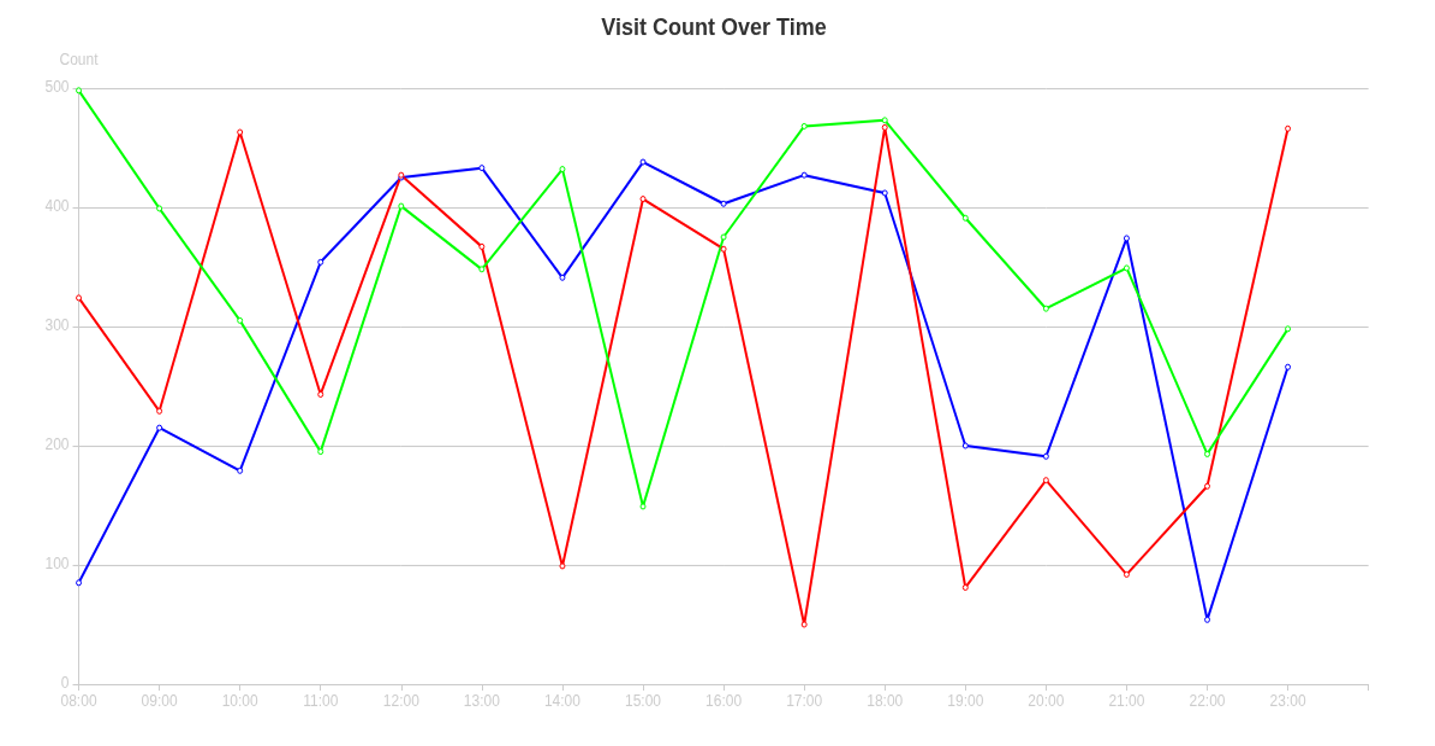} & \includegraphics[width=0.145\textwidth]{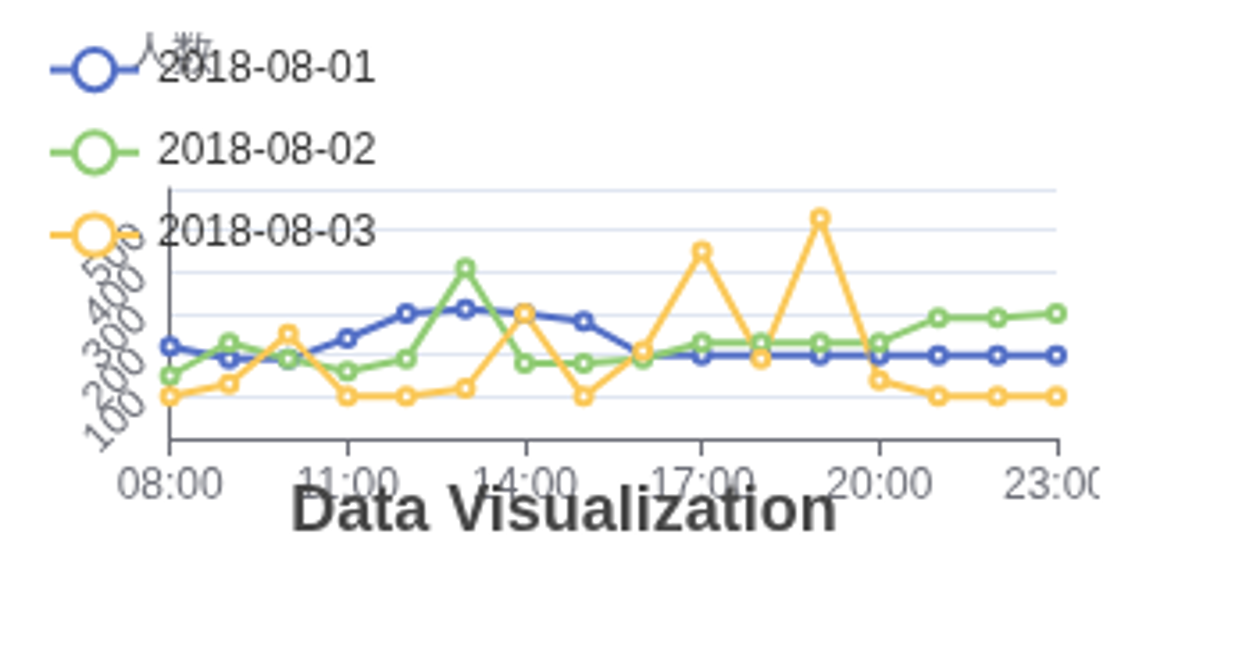} \\
    \end{tabular}
\end{adjustbox}
    
    \caption{Qualitative comparison of representative models across the six ChartInt tasks.}
    \label{fig:model_comparison}
    \Description{Grid of ground-truth and model-generated charts across six ChartInt tasks for representative models.}
\end{figure*}

\textbf{Data Semantic Degradation in Interactive Contexts.}
A sharp "Static-Dynamic Gap" emerges in the Interaction task. While the decline in code execution rates is negligible, Data Faithfulness scores plummet significantly. For instance, Gemini-3-Pro drops from \textbf{4.690} (Native Reconstruction) to \textbf{3.567} (Interaction).
This divergence reveals that current models struggle to maintain data integrity when burdened with interaction logic. Although they successfully render the visual components of an interaction (e.g., a tooltip box), they frequently fail to bind the correct data values to the specific event triggers. Consequently, the chart appears to function interactively but conveys erroneous information upon user action, indicating that the complexity of dynamic logic distracts the model from preserving fundamental data semantics.

\subsubsection{The Polarization of Interaction Costs.}
To investigate the mechanics of this degradation, we analyzed the correlation between the interaction evaluation score from GPT and the resulting drop in static performance ($\Delta = S_{static} - S_{interactive}$), as visualized in Figure \ref{fig:polarization_analysis}. The results reveal a distinct "Polarization Phenomenon" characterized by non-linear costs. While low-tier models (Scores 0-1) show negligible variance due to uniform failure, a significant "Valley of Collapse" emerges in the mid-range (Scores 2-3). In this regime, partial competence in dynamic logic destabilizes the chart structure, incurring the highest penalty on visual and data fidelity. Conversely, high-quality interactions (Scores 4-5) exhibit a "Synergy" effect characterized by a negative drop; here, the structural mastery required for perfect interactions inadvertently refines static semantics, correcting latent errors. This dichotomy highlights a critical Capability Threshold, indicating that MLLMs must surpass a specific proficiency level to support interactivity without compromising—and potentially enhancing—basic visualization integrity.

\subsubsection{Failure Analysis.}
Figure~\ref{fig:model_comparison} reveals divergent failure modes. Strong models (e.g., GPT-4o) exhibit \textit{Interaction-Induced Layout Collapse}: triggering a tooltip shifts the coordinate system downwards, causing axis labels to overlap with the legend. The expected interaction is present, but its dynamic state disrupts the static layout. Weaker models (e.g., InternVL3.5-2B) exhibit \textit{Total Functional Omission}, generating visually plausible static charts without the required interaction configurations. These outputs remain unresponsive despite appearing correct initially. This contrast suggests a progression from constructing interaction logic to integrating it without compromising structure. Reliable generation requires joint reasoning about event execution and the resulting visual state; initial static inspection can substantially overestimate interactive reliability.

\subsection{RQ2: Real-World Usability and the Reproduction-Adaptability Gap}

RQ2 investigates a critical question for practical deployment: Does proficiency in static reproduction translate to real-world usability? Beyond monotonic visual replication, real-world workflows require models to act as reliable assistants—updating data pipelines, tweaking aesthetics, and migrating styles without breaking existing functionality. We utilize the editing tasks to test this adaptability. 

Quantitative results in Table \ref{tab:finegrained} reveal a universal performance decline across all three editing tasks compared to native generation, confirming that there is still a significant gap between merely reproducing charts and meeting real-world requirements. In Style Adjustment, even the robust Gemini-3-Pro regresses, failing to preserve data labels present in the source (Figure \ref{fig:model_comparison}), while GPT-4o's grasp of data semantics deteriorates significantly under style constraints. In Data Update, despite the theoretical precision of programmatic file reading, the marked drop in Text metrics (reflecting legend alignment) and GPT-based Data scores reveals a critical inability to interpret and bind file-based semantics. The decline reaches its nadir in Style Transfer, where performance collapses across the board; even Gemini's metrics nearly halve, and GPT-4o suffers from "Semantic Collapse," incorrectly rendering a line chart as a bar chart because the visual reference style overrides the underlying data structure.

\subsection{Overall Performance Assessment}

We aggregate the metrics with a final score $S_{\text{F}}$. Before aggregation, GPT-based scores $s \in [1,5]$ are normalized to $[0,1]$ using $\operatorname{clip}((s-1)/4,0,1)$; OCR F1 and Color-IoU are used directly. Table~\ref{tab:finegrained} reports the original GPT scores.
\begin{equation}
S_{\text{F}} = R_{\text{exec}} \times
(0.3 S_{\text{data}} + 0.2 S_{\text{text}} + 0.2 S_{\text{vis}}
+ 0.15 S_{\text{style}} + 0.15 S_{\text{color}}).
\end{equation}
Here $R_{\text{exec}}$ gates non-executable outputs. Semantic terms receive 70\% of the quality weight, prioritizing information fidelity over appearance. Task-specific and dimension-specific results in Table~\ref{tab:finegrained} remain the primary evidence; the final score provides a complementary summary.

Weight perturbation confirms that conclusions do not depend on one tuned formula: keeping $R_{\text{exec}}$ fixed and sampling 1000 settings after $\pm$0.03/0.05/0.10/0.20 perturbations gives mean Spearman 0.9998/0.9994/0.9987/0.9976 and Kendall $\tau_b$ 0.9987/0.9971/0.9936/0.9876. Figure~\ref{fig:hot} reveals a stable difficulty hierarchy, \textit{Native}$\approx$\textit{Perturbed}$<$\textit{Interaction}$\approx$\textit{Style Adjustment}$<$\textit{Data Update}$<$\textit{Style Transfer}, marking the boundary between visual mimicry and reliable chart maintenance. While strong models (e.g., Gemini-3-Pro) maintain robustness in generation and basic interaction tasks, a sharp "performance cliff" emerges in high-abstraction maintenance tasks. Specifically, the significant score drops in Data Update and Style Transfer indicate that current MLLMs lack the compositional reasoning required to disentangle chart components, marking the boundary between simple visual mimicry and true intelligent adaptability.

\subsection{Ablation Study}
\label{sec:abl}
To balance evaluation robustness with computational efficiency, we perform ablation studies on the number of preset states evaluated per chart. Validating our state budget with a 50-chart ablation using five representative states per chart, we observe diminishing returns: the Fail@$k$ metrics for $k \in \{1, 2, 3, 5\}$ are 0.52, 0.52, 0.54, and 0.54, respectively. Furthermore, strong conditional correlations exist among state failures. If the first two core states both fail, the remaining three fail with a probability of 0.9615; conversely, if both succeed, the remaining three succeed with a probability of 0.9583. These results demonstrate that evaluating just one or two core states effectively captures the vast majority of model failures while keeping evaluation costs tractable for large-scale benchmarks.

\subsection{Cross-Validation Study of LLM Judges}
\label{sec:human}
While the LLM-as-Judge paradigm is highly scalable, it often raises concerns regarding inherent model biases. To investigate this, we perform a comprehensive set of cross-validation experiments. First, a 200-sample blind study involving five graduate-level raters demonstrates strong human agreement (ICC(2,1) = 0.793, 0.811, and 0.743 for logic, data, and style, respectively). Building on this human baseline, GPT-4o shows strong alignment with human consensus (MAE 0.540, sample-level Spearman 0.813, model-level Pearson/Spearman 0.959/0.961). Furthermore, to ensure our pipeline is not overly reliant on a single model, an independent audit using Claude Opus 4.6 confirms high agreement with both human raters (sample/model Spearman 0.8795/0.9541) and GPT-4o (sample Pearson/Spearman 0.754/0.748; model-level 0.944/0.938). Together, these validations confirm that our automated evaluation is statistically robust and representative of expert human judgment.

\section{Discussion \& future Work}

The results of ChartInt suggest that chart-to-code generation should be
viewed beyond static visual reconstruction, toward the generation of
executable and state-aware visualizations. Our experiments show that
high-quality chart generation requires models to jointly preserve visual
appearance, data semantics, interaction logic, and functionality across
different visualization states. In particular, the observed static-dynamic
gap and the performance degradation on real-world editing tasks indicate
that future chart-oriented MLLMs should move from one-shot code generation
toward more structured reasoning over visualization states and program
behavior. Promising directions include interaction-aware training, explicit event--state modeling, tool use, and iterative refinement\cite{zhang2026generative,zhang2026omniverifier,liu2026vidforensics}. Future work can explore multi-step editing, richer cross-chart transformations, and agents that inspect, modify, execute, and verify charts through successive interaction cycles. Beyond generating runnable scripts, models need to maintain stable bindings between raw data and visual encodings during state transitions, so that interactive operations do not corrupt semantic mappings. Meanwhile, evaluation paradigms should complement static visual metrics with functional tests to rigorously assess the interactive reliability of chart-generation models.

\section{Conclusion}
\label{con}
We introduce ChartInt, the first benchmark specifically designed to evaluate multimodal large language models on interactive chart-to-code generation, addressing the critical limitation of existing benchmarks that focus solely on static visual similarity. By proposing the Preset State Snapshot Protocol, we transform dynamic interaction behaviors into reproducible static evaluation targets, enabling a rigorous assessment of models across interaction verification and real-world editing tasks like data updates and style transfers. Our experiments reveal a significant "Static-Dynamic Gap," where models capable of static reconstruction struggle to maintain data semantics during interactions, and a "Reproduction-Adaptability Gap" in real-world editing requirements, highlighting the urgent need for MLLMs to improve their compositional reasoning and dynamic logic capabilities for practical web visualization.

\section*{Ethical Considerations}

This work focuses on benchmarking multimodal large language models for
chart-to-code generation, interactive visualization, and visualization
editing. The benchmark is constructed from visualization examples and
programmatic tasks, and does not require sensitive personal information as benchmark input. Human participation is limited to the rating study described in Section~\ref{sec:human}. The
evaluation is performed on visualization code, rendered charts, and
predefined interaction states, without exposing personally identifiable
information. We use publicly available visualization examples as sources
for benchmark construction and do not intentionally include private or
confidential user data.
In particular, our
evaluation emphasizes executable correctness, data faithfulness, and
visual consistency, which can help identify cases where a visually
plausible chart contains incorrect or misleading information. We respect
the licensing and attribution requirements of the underlying
visualization libraries and publicly available resources used in
benchmark construction, and encourage future users of the benchmark to
follow the corresponding terms of use and attribution requirements.

\bibliographystyle{ACM-Reference-Format}
\bibliography{tmybib}

\end{document}